\documentclass[review]{elsarticle}

\usepackage{enumerate}
\usepackage{natbib}
\usepackage{tabularx}
\usepackage{enumitem}
\usepackage{subfigure}
\usepackage{makecell}
\usepackage[flushleft]{threeparttable}
\usepackage{algorithmic}
\usepackage{amsmath}

\usepackage{graphicx}
\usepackage[margin=1in,letterpaper]{geometry} %
\usepackage{multirow}
\usepackage[ruled]{algorithm2e}
\usepackage{xcolor}
\def\HiLi{\leavevmode\rlap{\hbox to \hsize{\color{yellow!50}\leaders\hrule height .8\baselineskip depth .5ex\hfill}}}

\journal{Journal of \LaTeX\ Templates}

\begin{document}

\begin{frontmatter}

\title{Evolutionary Optimization for Proactive and Dynamic Computing Resource Allocation in Open Radio Access Network}
%
\author[myaddress]{Gan Ruan}\ead{GXR847@cs.bham.ac.uk}

\author[myaddress]{Leandro L. Minku}
\ead{L.L.Minku@cs.bham.ac.uk}
\author[mysecondaryaddress]{Zhao Xu}\ead{zhao.xu@neclab.eu}
\author[myaddress,mythirdaddress]{Xin Yao}\ead{X.Yao@cs.bham.ac.uk}

\address[myaddress]{CERCIA, School of Computer Science, University of Birmingham, Edgbaston Birmingham B15 2TT, UK}

\address[mysecondaryaddress]{NEC Laboratories Europe, 69115 Heidelberg, Germany}
\address[mythirdaddress]{Department of Computer Science and Engineering, Southern University of Science and Technology, Shenzhen, China.}

\begin{abstract}

Intelligent techniques are urged to achieve automatic allocation of the computing resource in Open Radio Access Network (O-RAN), to save computing resource, increase utilization rate of them and decrease the delay. However, the existing problem formulation to solve this resource allocation problem is unsuitable as it defines the capacity utility of resource in an inappropriate way and tends to cause much delay. Moreover, the existing problem has only been attempted to be solved based on greedy search, which is not ideal as it could get stuck into local optima. Considering those, a new formulation that better describes the problem is proposed. In addition, as a well-known global search meta heuristic approach, an evolutionary algorithm (EA) is designed tailored for solving the new problem formulation, to find a resource allocation scheme to proactively and dynamically deploy the computing resource for processing upcoming traffic data. Experimental studies carried out on several real-world datasets and newly generated artificial datasets with more properties beyond the real-world datasets have demonstrated the significant superiority over a baseline greedy algorithm under different parameter settings. Moreover, experimental studies are taken to compare the proposed EA and two variants, to indicate the impact of different algorithm choices.
\end{abstract}

\end{frontmatter}


\section{Introduction}

Open Radio Access Network (O-RAN) is a newly proposed wireless network architecture to fulfill all kinds of demands posed by various applications, services and deployment of different operators and vendors, in the 5G or beyond 5G wireless systems \cite{singh2020evolution}. Therefore, O-RAN is emerging with two fundamental aspects: Openness and Intelligence \cite{gavrilovska2020cloud}. It provides open interfaces for different operators and vendors to deploy their own infrastructures and offer tailored services. In such way, the network becomes more complex and heavy than any networks before. It is therefore impossible for operators and vendors to depend on human intensive means of optimizing, deploying and operating the mobile networks. Therefore, novel intelligent technologies are welcome to assist the automation of the system \cite{gavrilovska2020cloud}.

One of the problems that requires the support of intelligent algorithms is resources management or allocation in O-RAN. An inappropriate allocation of resources could lead to high operational costs. O-RAN, as a type of RAN, also involves two major components, i.e. remote radio head (RRH) and baseband unit (BBU) due to its functionality of managing radio resources and processing signal data. One of the most challenging problems in O-RAN is how to allocate the computing resources (BBUs) to different RRHs such that they can handle traffic data produced by applications and software in the network. Furthermore, the vendors and operators need to save the computing resources to decrease their capital expenditures and deployment cost. At the same time, they should also be able to provide competitive Internet services to users in the network such that they have good user experience, which requires less delay in the network everyday. Therefore, it is a significant problem how to proactively allocate the computing resources in the network to achieve the optimization goals, i.e. decreasing required number of computing units and increasing the resources utilization rate as well as decreasing the delay.


In order to address this problem, a problem formulation has been proposed in \cite{chen2018deep}. Besides, a Multivariate Long Short-Term Memory (MuLSTM) model is put forward to predict the traffic data in a future period for each RRH in the network. Lastly, a greedy algorithm was suggested to find a solution for this problem based on the predicted traffic data. Specifically, the idea is to cluster RRHs with complementary traffic pattern to a BBU, such that the BBU capacity can be shared among those RRHs at different hours of a day. Therefore, two metrics have been proposed to measure the complementarity of each cluster, which includes the peak distribution and the so-called capacity utility. Peak distribution measures how the peak hour of each RRH in the cluster widely distributes in different hours of a day, so that the BBU capacity can be fully occupied by the RRHs. Capacity utility aims to make the aggregated traffic of all RRHs in each cluster close to the BBU capacity.    After defining this, the greedy algorithm iteratively connects a random RRH to its closest cluster and select the cluster that gets the best fitness value until the termination condition is satisfied, where the fitness function is calculated on the pre-defined complementarity of clusters.

However, there are several limitations in the framework of the existing work \cite{chen2018deep}, both regarding the problem definition and the algorithm. Specifically, the peak distribution in the formulation is needless as it is not related to the optimizing objectives which are decreasing required number of computing units and increasing the resources utilization rate as well as decreasing the delay. The reason will be analyzed in Section \ref{sec:newProb}. Besides, the existing problem formulation is defined as making the averaged total traffic data of each cluster over a 24 hours period close to the computing resource capacity when it is optimized. This may result in cases where the solution may be unable to cope with high traffic in some hours of the day, while being underutilized in other hours of the day when the traffic may be lower. Moreover, when optimizing the existing formulation, solutions with more delay in the network have more chance to survive than those with less computing unit utilization rate. Therefore, the algorithm is likely to be biased towards solutions where the network will present delays. In addition, greedy algorithms are known to easily get stuck into local optima \cite{simmons2019beware} \cite{gao2018randomized}.



To overcome these problems, the following research questions are answered in this paper:
\begin{enumerate}
\item How to appropriately formulate the computing resource allocation problem?
\item How to design an evolutionary algorithm tailored for solving the new formulation?
\begin{itemize}
\item Does the proposed EA outperform a greedy algorithm? Under what conditions?
\item What is the influence of different algorithm's design choices on its performance?
\end{itemize}
\end{enumerate}

To answer the first research question, after considering the above-mentioned weaknesses of the existing formulation, a new problem formulation is firstly proposed that does not have the weaknesses of the existing formulation. Specifically, the proposed problem formulation tries to make the total traffic data in each cluster close to the computing resource capacity at each hour of a day when it is optimized, so that an optimal solution would not lead to delays in certain periods of the day while other periods may have resources underutilized. In addition, it equally considers the effect of the delay and computing resource utilization rate on the fitness value of solutions. Besides, the new problem formulation removes the needless metric in the existing formulation. A mathematical analysis of the reasons why the proposed problem formulation is more adequate than the existing one will be presented. Evolutionary algorithm, as an algorithm with good ability of global search \cite{eiben2003introduction}, is an alternative to be selected as the algorithm to solve the new problem formulation. However, existing evolutionary algorithms are not suitable for the computing resource allocation problem in this paper as their representation and evolutionary operators are not directly applicable. Therefore, we propose an evolutionary algorithm (EA) tailored for solving the resource allocation problem in this paper. Specifically, it mainly includes initialization, mutation operator and random cluster splitting. Initialization aims to randomly generate an initialized population with feasible solutions on the problem representation. Mutation operator tries to produce feasible offspring solutions. Suppose those RRHs assigned with the same computing resource (BBU) are in one cluster, a solution for the problem is a set of clusters. Random cluster splitting aims to randomly select a cluster from a solution in the optimized population of the previous day and randomly split it into two clusters, so as to produce an initial population for the problem of the next day. The proposed evolutionary algorithm is named as SplitEA. The SplitEA uses a multivariate long short-term memory model to predict the traffic data in a future period of time. This enables the algorithm to capture the temporal dependency and spatial correlation among base station traffic patterns \cite{chen2018deep}. Experimental studies have been carried out on a series of real-world and artificial datasets to answer two sub-questions of the second research question, which successfully validates the effectiveness of the proposed evolutionary algorithm over the greedy algorithm.

After answering those research questions, we can conclude the novel contributions of this paper as follows:
\begin{enumerate}
\item A novel problem formulation that better describes the computing resource allocation problem is proposed;
\item An evolutionary algorithm is specifically designed to solve the new problem formulation;
\begin{itemize}
\item Comprehensive experimental studies have been conducted to verify the proposed algorithm on various  datasets, showing that the proposed evolutionary algorithm performs significantly better than the greedy algorithm on real-world datasets and other different scenarios with different points distribution and traffic pattern.
\item A performance analysis of our proposed algorithm on different design choices shows the effectiveness of the random cluster splitting, through comparing SplitEA and its two variants (CopyEA and RandEA) on most real-world and artificial datasets. CopyEA replaces the random cluster splitting of SplitEA with copying all solutions of the problem at the previous day as the initial population for the next day, while RandEA leverages the proposed initialization for problems at all days.
\end{itemize}
\end{enumerate}
The rest of this paper is organized as follows: Section 2 introduces the background on the computing resource allocation problem in O-RAN and general approaches to resources allocation problems. Section 3 presented the proposed problem formulation. Section 4 describes the details of the proposed evolutionary algorithm. Experimental studies are presented in Section 5. Section 6 summarizes this paper and gives an outlook of the future work.

\section{Background}

\subsection{Related Work on the Computing Resource Allocation Problem}

Considering that the computing resources are required to be deployed beforehand to tackle the future coming data in O-RAN, it is therefore a significant problem how to proactively allocate the computing resources
in the network to decrease required number of BBUs and increase the resources utilization rate as well as decrease the delay. In order to solve this problem, it is firstly required to predict the traffic data of a set of RRHs in the network. The authors in \cite{chen2018deep} designed a sequence to sequence model that uses unified multivariate LSTM model. The model receives the traffic data of all RRHs in the network at 24 hours of the previous day as the input and outputs the traffic data of all RRHs at 24 hours of the next day. The authors in \cite{chen2018deep} also created a problem formulation. Specifically, the authors considers to cluster RRHs with complementary traffic patterns together and assign each cluster with one BBU, such that the BBU capacity can be shared by those clustered RRHs. To define the complementarity of RRHs, the authors considered two aspects: peak distribution and so-called capacity utility, as introduced in \cite{chen2018deep}.

Peak distribution: In order to achieve the goal that BBU capacity can be shared among those RRHs clustered to the BBU, the peak traffic volume of those RRHs should be scattered at different hours of a day. Therefore, inspired from the idea of entropy, the authors in \cite{chen2018deep} developed an indicator to measure the peak distribution of those clustered RRHs. To make the problem formulation more mathematical, suppose points represent RRHs and those RRHs assigned with the same BBU are in one cluster. Specifically, given a set of clustered $n$ points $C = (r_1,...,r_i,...,r_n)$, peak hours for each point in $C$ are defined as follows, which is calculated based on the predicted traffic volumes for the next day, 
\begin{equation}
T(r_i)=\{t_{i_1},t_{i_2},...,t_{i_m}\},  1\leq i_m \leq 24
\end{equation}
where $t_{i_m}$ denotes the $m_{th}$ peak time of $r_i$. Then the Shannon entropy of the peak hours of the set of clustered points $T(C)=\cup T(r_i)$ is calculated as follows:
\begin{equation}
H(C)=-\sum^{J}_{j=1}p_jlog(p_j)
\end{equation}
where $J=|T(C)|$ is the total quantity of peaks in $C$ and $p_k$ is the probability of observing the corresponding peak hour in the set $T(C)$. A larger entropy value of a cluster indicates that the points are more complementary in the cluster w.r.t. traffic patterns.

\begin{figure}[htp]
    \centering
    \includegraphics[width=1\linewidth]{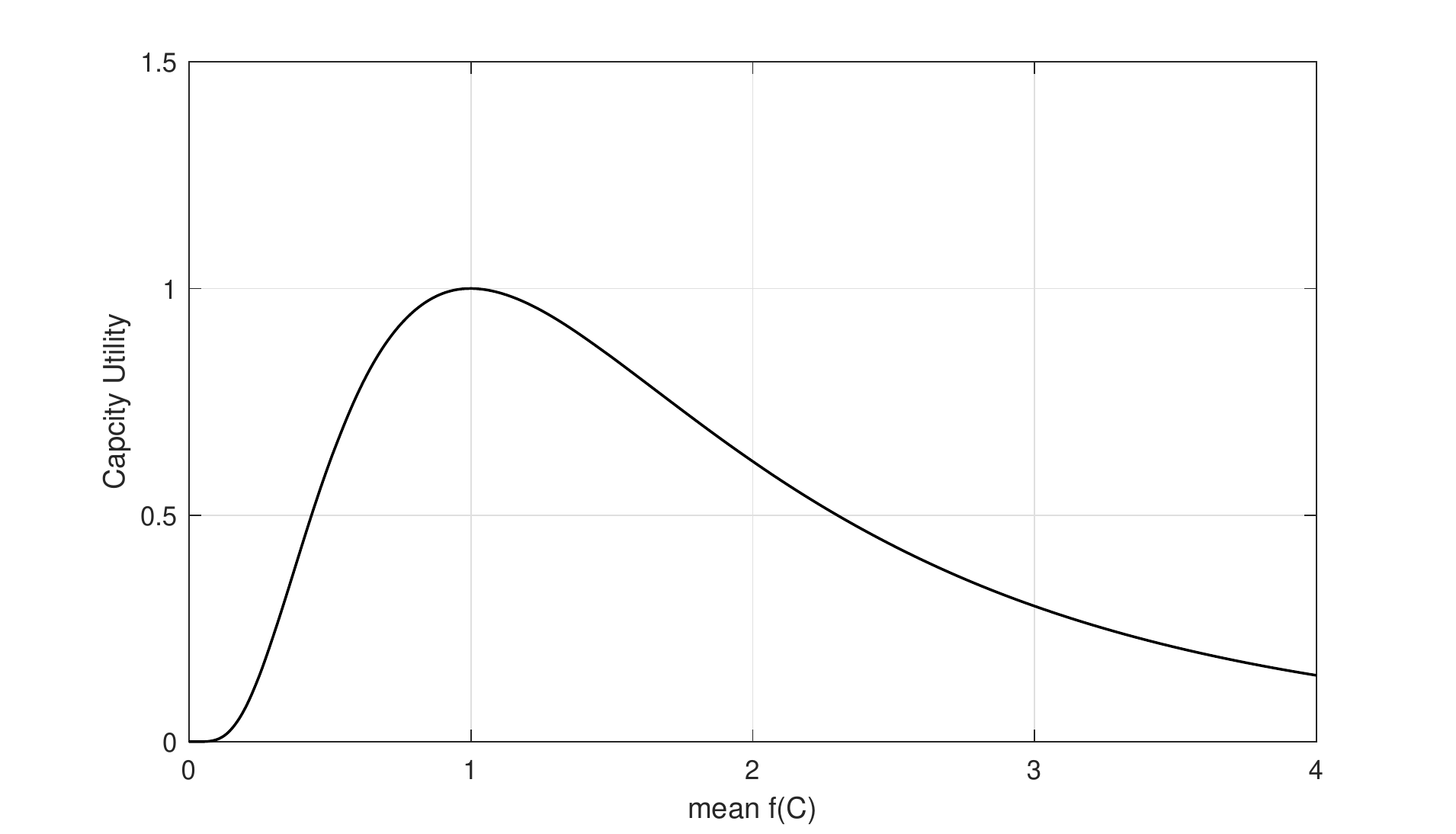}
\caption{The curve of the so-called capacity utility function in \cite{chen2018deep}, which reaches its
maximum when the cluster traffic volume equals 1.}
\label{ExistingCapaUtility}
\end{figure}

To make full use of the BBU mapped to a cluster C, the aggregated cluster traffic should be close to the BBU capacity in different hours of the day. Meanwhile, to prevent the BBU from overload, the aggregated cluster traffic should not exceed the BBU capacity too much. 
To describe the problem formulation in a more concise way, we assume the utility of BBU capacity is 1. In the existing work \cite{chen2018deep}, the so-called capacity utility is defined as follows,
\begin{equation}
U(C)=[mean\; \textbf{f}(C)]^{-ln({mean\; \textbf{f}(C)})}
\end{equation}
where $\textbf{f}(C) = \sum^n_{i=1}\textbf{f}(r_i) = \bigg(\sum^n_{i=1}f_1(r_i),..., \sum^n_{i=1}f_h(r_i),..., \sum^n_{i=1}f_{24}(r_i) \bigg)$ is the aggregated traffic volume of the cluster $C$. Therefore, $mean\; \textbf{f}(C) = \frac{1}{24}\sum^{24}_{h=1} \sum^n_{i=1}f_h(r_i)$.
Figure \ref{ExistingCapaUtility} displays the curve of the so-called capacity utility function, which achieves its maximum when the mean aggregated traffic volume is equal to the BBU capacity 1. Therefore, the complementarity of the RRHs cluster $C$ is calculated as follows \cite{chen2018deep},
\begin{equation}
M(C)=U(C)*H(C)
=-[mean f(C)]^{-ln({mean f(C)})}\sum^{J}_{j=1}p_jlog(p_j)
\end{equation}

To solve this problem formulation, the authors propose  a Distance-Constrained Complementarity-Aware algorithm to cluster close RRHs to the same BBU. Note that quality-of-service requirements may be affected by the propagation delay as the distance between RRHs and BBU increases. Besides, there may be also the communication delay among RRHs when enabling machine to machine communications such as handover \cite{1991Handover} in the mobile network. Therefore, to alleviate the delay in the network, when allocating the computing resources in the network, the distance between any two RRHs assigned to the same BBU should constrained within a range. The basic idea of the algorithm is to iteratively cluster close RRHs into groups, such that the complementarity of RRHs in all clusters is maximized. Specifically, the authors design a fitness function considering the complementarity of RRHs and the distance of any two RRHs in the cluster. The algorithm  greedily assigns randomly selected RRHs to the adjacent cluster with highest fitness value until the termination criteria is satisfied. More details about the algorithm can be found in \cite{chen2018deep}.

However, there are several limitations in the existing framework \cite{chen2018deep}. Firstly, in the definition of so-called capacity utility, there are two main limitations. It is clear from Figure \ref{ExistingCapaUtility} that the so-called capacity utility is maximum when $mean\; \textbf{f}(C)$ reaches BBU capacity 1. However, $mean\; \textbf{f}(C)$ is the mean traffic volume of all RRHs in this cluster under 24 hours. It is possible that a clustering scheme causing much delay at some hours while very low BBU occupation rate at other hours may be still a good one. Besides, according to the curve, when $mean\; \textbf{f}(C)$ is extremely large, it still gets the same capacity utility as that with BBU almost unoccupied. To this end, the clustering scheme found by this problem formulation may bias to the cases causing much delay. Another limitation of the existing problem formulation is that the peak distribution is needless as this definition does not have a direct relationship to the optimization goals. A simple example will be presented to state why it is not required in Section \ref{sec:newProb}. Besides the limitations in the problem formulation, the used greedy algorithm is known to easily get stuck in the local optima \cite{simmons2019beware} \cite{gao2018randomized}. The reason is that the idea of greedy algorithms is to greedily choose the currently best solution to survive at each iteration.

\subsection{Time Series Prediction Methods}

Given that in the computing resource problem traffic data needs to be predicted beforehand for the proactive allocation of the computing resource, several commonly used time series prediction models in the literature \cite{sang2002predictability} \cite{box2015time} \cite{chen2018deep} are presented.

Autoregressive integrated moving Average (ARIMA) is one of the most widely used time series analyses model and has been successfully applied to solve many short-term forecasting problems \cite{sang2002predictability}. However, it gets worse prediction errors and confidence on long-term forecasting problems where multiple future steps need to be predicted \cite{box2015time}. In addition, Artificial Neural Network (ANN) models are also leveraged to learn time series properties and predict the trend of the data in the future through using a
sliding-window-based strategy \cite{dorffner1996neural}, which has been used in many domains like operation research \cite{zhang2005neural} and financial market \cite{azoff1994neural}. However, it is difficult for ANNs to model the temporal dependency between the elements in each time series window as analyzed in \cite{chen2018deep}. Moreover, a multivariate LSTM is proposed in \cite{chen2018deep} to learn the temporal dependency and spatial correlation of RRH traffic data.

\subsection{Existing Optimization Algorithms for Resource Allocation Problems}

There are three types of methods for solving resource allocation problems in the literature, which are linear programming \cite{moens2014vnf} \cite{gupta2015service}, heuristic and metaheuristic methods. All of them have their advantages and disadvantages. Linear programming tends to find an exact solution for resource allocation problems \cite{moens2014vnf}. However, considering that most problems are highly complex, nonlinear and with constraints, people always simplify the modeling of the linear programming, thus results in inaccurate solutions for the problem. As for a kind of heuristic algorithm, greedy methods are a representative approach to solve the resource allocation problem \cite{riggio2015virtual} \cite{mijumbi2015design}\cite{khan2019hybrid}. However, it is easy for the greedy algorithm to get stuck into local optima. Many metaheuristic algorithms like evolutionary approaches have been widely applied into addressing the resource allocation problem due to its competent global search ability \cite{perveen2021dynamic} \cite{robinson2011market} \cite{lewis2009evolutionary} \cite{lewis2010resource} \cite{wang2010multi} \cite{lewis2008evolutionary} \cite{salcedo2008optimal} \cite{salcedo2008assignment}.

Given that the problem we deal with in this paper is non-linear and has complex objective function and constraint, it is unable for any linear programming methods to solve. If the linear programming is used to solve this problem, the problem formulation needs to be simplified, thus resulting inaccurate solution for the problem, which is not a good trial. In addition, greedy algorithms, as a heuristic algorithm, easily get stuck into the local optimal, which has been verified and stated in \cite{simmons2019beware} \cite{gao2018randomized}. Therefore, the greedy algorithm that has been used in the existing work \cite{chen2018deep} may prevents the search of better solutions, which is not ideal Moreover, most existing evolutionary algorithms are not suitable for the computing resource allocation problem in this paper. The reason is that the resource allocation problem formulation in this paper is different from those in the literature \cite{mousa2017k}, which may cause the differences in the processes of the evolutionary algorithm including solution representation, population initialization, genetic operators and constraint handling \cite{mousa2017k}. For example, different people may represent a solution with a vector \cite{salcedo2004hybrid}, a matrix  \cite{chen2018deep} \cite{salcedo2004hybrid} or even a graph \cite{mousa2017k}. It is easy to understand that genetic operators will be different if the solution representation is different. 

\section{Proposed Formulation of the Computing Resource Allocation Problem}
\label{sec:newProb}

The limitations of the existing problem formulation are stated as follows. The peak distribution in the formulation is redundant as optimizing the defined problem with the peak distribution does not directly optimizing the three objectives. Besides, the existing problem formulation is defined as making the averaged total traffic data in each cluster of 24 hours close to the computing resource capacity when it is optimized. Moreover, when optimizing the existing formulation, solutions with more delay in the network have more chance to survive than those with less computing unit utilization rate.

Therefore, in this section, a new problem formulation is proposed to remedy the weaknesses of the existing problem formulation, which is trying to answer the first research question: How to appropriately formulate the computing resource allocation problem? To make the problem formulation more mathematical, points represent RRHs and the data of each point means the traffic data of each RRH. In addition, assume those RRHs assigned with the same BBU are in one cluster. Suppose there are $N$ points $R = (r_1,...,r_i,...,r_N)$; each point $r_i$ has a fixed position $(r_i^1, r_i^2)$ (where $r_i^1$ and $r_i^2$ are the longitude and latitude, respectively) and traffic data at 24 hour of a day $\textit{\textbf{f}} = \big( f_0(r_i),..., f_h(r_i),...,f_{23}(r_i) \big)^T$, where $f_h(r_i)$ is sum of traffic data of $r_i$ from $h$-th to $(h+1)$-th hour at the current day.

At the end of each day, the clustering scheme of the next day needs to be found to deploy the BBUs to the RRHs in the network. Considering this background, the aim is to proactively cluster these $N$ points to $K$ clusters to achieve the optimization objective.  Therefore, this problem needs to firstly predict the traffic data of all points and then dynamically find an optimal solution through an optimization algorithm based on the predicted traffic data, such that BBU utilization rate is maximum, the delay in the network and the required number of BBUs is minimal. This problem can be also regarded as a time series clustering problem.

There may be communication delay among RRHs when enabling machine to machine communications such as handover \cite{1991Handover} in the mobile network. Therefore, to alleviate the delay in the network, when allocating the computing resources to the base stations in the network, the distance between any two RRHs assigned to the same BBU should be constrained within a range.

Let $\textit{\textbf{X}}=\{x_1, x_2,...,x_N\}$ be a solution vector where all points have been grouped into certain clusters. Let $K$ is the maximum number of clusters and $x_i=l$ if the $i$-th point is in the $l$-th cluster. Therefore, the range of $x_i=l$ is $[1,K]$. It is clear that at least one $x_i$ is equal to any value of $[1,K]$. In other words, there is at least one point in all clusters from 1 to $K$. The objective function is presented as follows:
\begin{equation}\label{eq:def of problem}
\begin{split}
\begin{aligned}[t]
\begin{cases}
    \min \mathbf{F}(\mathbf{X})= w*K + \frac{1}{K}\sum\limits_{k=1}^{K}U(C_k) \\
    s.t.~dist(r_u, r_v)\leq \tau ~(\forall ~r_u, r_v \in C_k)
\end{cases}
\end{aligned}
\end{split}
\end{equation}
where $K = max(x_i)~(i = 1, 2,..., N$) is the number of clusters; $w \in (0,1]$ is a parameter that controls the weight of $K$ and $\frac{1}{K}\sum\limits_{k=1}^{K}U(C_k)$; $dist(r_u, r_v)$ is the distance of any two points $r_u$ and $r_v$ in $k$-th cluster $C_k$; $\tau$ is a threshold controlling the distance of neighboring points; $U(C_k)$ is the difference between the sum data of points and 1 in the $k$-th cluster, which is defined as:
\begin{equation}\label{eq:def of uck}
U(C_k) = \frac{1}{24}*\sum\limits_{h=0}^{23}|f_h(C_k)-1|;
\end{equation}
where $f_h(C_k)$ is the sum traffic data of all points in $k$-th cluster, which is defined as:
\begin{equation}
f_h(C_k) = \sum\limits_{r_m \in C_k}f_h(r_m)
\end{equation}
where $C_k$ is the set of all point in the $k$-th cluster, which is presented as follows: 
\begin{equation}
C_k = \{r_m|x_m = k\}
\end{equation}

\textbf{Remark 1:} The reason why we add the $K$ as one part in the fitness function is that the number of clusters is one of the main goal to be minimized to decrease the capital investment of operators and vendors. $w$ is a parameter controlling the weight of number of clusters and the summation of delay and BBU capacity, which gives operators and vendors a choice to make a balance between the investment and the user experience such that they can set it according to their own spirit.


\begin{figure}[htp]
    \centering
    \includegraphics[width=0.95\linewidth]{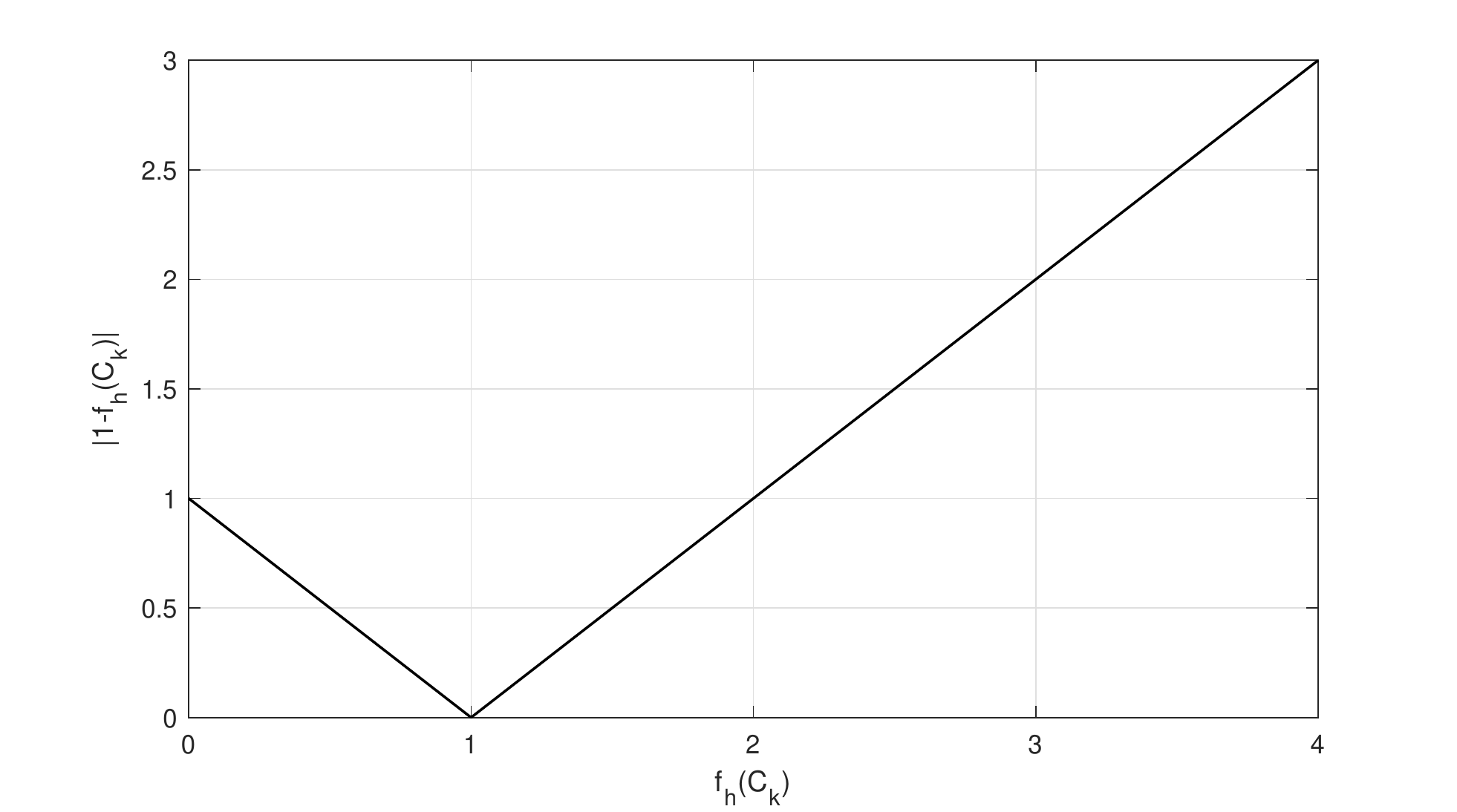}
\caption{The curve of $|1-f_h(C_k)|$ function in equation \ref{eq:def of uck}, which reaches its
minimal value when the clustered traffic volume equals 1.}
\label{fig:ImprovedUCk}
\end{figure}
\textbf{Remark 2:} The $U(C_k)$ in equation (\ref{eq:def of uck}) makes more sense than the existing one. The reasons are two-fold. Firstly, minimizing equation (\ref{eq:def of uck}) is trying to make the summed data of each cluster close to the BBU capacity of 1 at 24 hours of a day rather than the existing one where the average summed data of 24 hours in each cluster is close to 1. In addition, the curve of $y = |f_h(C_k)-1|$ is shown in Figure \ref{fig:ImprovedUCk}. It can be seen that when minimizing equation (\ref{eq:def of uck}), solutions with underutilized BBU and those with delay are given the same chance and those with much delay are given large fitness value.

\begin{table}[ht!]
\caption{The values of $1-U(C_k)$, $H(C_k)$ and their mean value of different clustering schemes for different datasets to state why the $H(C_k)$ is unnecessary.}
\label{table:exampleHck}
\scriptsize
\begin{center}
\begin{tabular}[width=\linewidth]{|c|c|cc|cc|cc|cc|c|}
  \hline
  Datasets & Clustering &   1-UC1 & HC1 & 1-UC2 & HC2 & 1-UC3 & HC3 & mean(1-U) & meanM  \\
  \hline
  \multirow{5}{*}{\makecell{{1h,  2h,  3h} \\ {[\textbf{0.8}, 0.5, 0.3]}\\{[0.2, \textbf{0.7}, 0.1]}\\{[0.2, 0.6, \textbf{0.7}]}}} & 12, 3  & 0.733 &  1 &  0.5  & 0 & NULL & NULL & 0.617 & 0.367\\
  & 13, 2 & 0.967 &  1 &  0.333  & 0 & NULL & NULL & \textbf{0.65} & 0.483\\
  & 1, 23 & 0.533 &  0 &  0.633  & 1 & NULL & NULL & 0.583 & 0.317\\
  & 1, 2, 3 & 0.533 &  0 &  0.333  & 0 & 0.5 & 0 & 0.456 & 0\\
  & 123 & 0.633 &  1.585 &  NULL  & NULL & NULL & NULL & 0.633 & \textbf{1.004}\\
  \hline  
  \multirow{5}{*}{\makecell{{1h,  2h,  3h} \\ {[\textbf{0.8}, 0.5, 0.3]}\\{[\textbf{0.7}, 0.2, 0.1]}\\{[0.2, 0.6, \textbf{0.7}]}}} & 12, 3  & 0.533 &  0 &  0.5  & 0 & NULL & NULL & 0.517 & 0\\
  & 13, 2 & 0.967 &  1 &  0.333  & 0 & NULL & NULL & 0.65 & 0.483\\
  & 1, 23 & 0.533 &  0 &  0.833  & 1 & NULL & NULL & \textbf{0.683} & 0.417\\
  & 1, 2, 3 & 0.533 &  0 &  0.333  & 0 & 0.5 & 0 & 0.456 & 0\\
  & 123 & 0.633 &  0.918 &  NULL  & NULL & NULL & NULL & 0.633 & \textbf{0.581}\\
  \hline  
  \multirow{5}{*}{\makecell{{1h,  2h,  3h} \\ {[\textbf{0.8}, 0.5, 0.3]}\\{[\textbf{0.7}, 0.2, 0.1]}\\{[\textbf{0.7}, 0.6, 0.2]}}} & 12, 3  & 0.533 &  0 &  0.5  & 0 & NULL & NULL & 0.517 & 0\\
  & 13, 2 & 0.633 &  0 &  0.333  & 0 & NULL & NULL & 0.483 & 0\\
  & 1, 23 & 0.533 &  0 &  0.567  & 1 & NULL & NULL & \textbf{0.55} & 0\\
  & 1, 2, 3 & 0.533 &  0 &  0.333  & 0 & 0.5 & 0 & 0.456 & 0\\
  & 123 & 0.367 &  0 &  NULL  & NULL & NULL & NULL & 0.367 & 0\\
  \hline
  \hline
  \multirow{5}{*}{\makecell{{1h,  2h,  3h} \\ {[\textbf{0.18}, 0.15, 0.13]}\\{[0.12, \textbf{0.17}, 0.11]}\\{[0.12, 0.16, \textbf{0.17}]}}} & 12, 3  & 0.287 &  1 &  0.15  & 0 & NULL & NULL & 0.218 & 0.143\\
  & 13, 2 & 0.303 &  1 &  0.133  & 0 & NULL & NULL & 0.218 & 0.152\\
  & 1, 23 & 0.153 &  0 &  0.283  & 1 & NULL & NULL & 0.218 & 0.142\\
  & 1, 2, 3 & 0.153 &  0 &  0.133  & 0 & 0.15 & 0 & 0.146 & 0\\
  & 123 & 0.437 &  1.585 &  NULL  & NULL & NULL & NULL & \textbf{0.437} & \textbf{0.692}\\
  \hline
  \multirow{5}{*}{\makecell{{1h,  2h,  3h} \\ {[\textbf{0.18}, 0.15, 0.13]}\\{[\textbf{0.17}, 0.12, 0.11]}\\{[0.12, 0.16, \textbf{0.17}]}}} & 12, 3  & 0.287 &  0 &  0.15  & 0 & NULL & NULL & 0.218 & 0\\
  & 13, 2 & 0.303 &  1 &  0.133  & 0 & NULL & NULL & 0.218 & 0.152\\
  & 1, 23 & 0.153 &  0 &  0.283  & 1 & NULL & NULL & 0.218 & 0.142\\
  & 1, 2, 3 & 0.153 &  0 &  0.133  & 0 & 0.15 & 0 & 0.146 & 0\\
  & 123 & 0.437 &  0.918 &  NULL  & NULL & NULL & NULL & \textbf{0.437} & \textbf{0.401}\\
  \hline
  \multirow{5}{*}{\makecell{{1h,  2h,  3h} \\ {[\textbf{0.18}, 0.15, 0.13]}\\{[\textbf{0.17}, 0.12, 0.11]}\\{[\textbf{0.17}, 0.16, 0.12]}}} & 12, 3  & 0.287 &  0 &  0.15  & 0 & NULL & NULL & 0.218 & 0\\
  & 13, 2 & 0.303 &  0 &  0.133  & 0 & NULL & NULL & 0.218 & 0\\
  & 1, 23 & 0.283 &  0 &  0.153  & 0 & NULL & NULL & 0.218 & 0\\
  & 1, 2, 3 & 0.153 &  0 &  0.133  & 0 & 0.15 & 0 & 0.146 & 0\\
  & 123 & 0.437 &  0 &  NULL  & NULL & NULL & NULL & \textbf{0.437} & 0\\
  \hline  
\end{tabular}
\end{center}
\vspace{-0.5cm}
\begin{tablenotes}
      \normalsize
      \item In the column of `Datasets', each line represents the traffic data of each point at three hour, in which the peak traffic value is highlighted in bold face. In the column of `Clustering', points with the numbers divided by the comma are in the same cluster. For example, `12,3' means first and second point are in the same cluster while the third point is an isolated cluster. In the last column, the best values obtained by the corresponding clustering scheme are highlighted in bold face. There are five different clustering choices shown in the second column. The clustering scheme with the largest values selected by $mean(1-U)$ or $meanM$ is also highlighted with bold font. 
    \end{tablenotes}
\end{table}

\textbf{Remark 3:} Why we do not need the peak distribution $H(C)$ in the fitness function? It is because optimizing $H(C)$ does not have a direct relation with three above-mentioned optimization goals. A simple example with several points and their traffic data at several hours is given to explain the reason. In this example, several datasets with different cases are generated, each of which has three points to be clustered and each point have the traffic data at three hours to make $U(C_k)$ and $H(C_k)$ easily calculated. As the $U(C_k)$ in the fitness function is to be minimized and $H(C_k)$ is to be maximized, let $M(C_k) = (1 - U(C_k)) * H(C_k)$ to be maximized. Therefore, the clustering that gets the largest value is the best one.

The clustering schemes and the values of $1-U(C_k)$, $H(C_k)$ and the mean of them is shown in Table \ref{table:exampleHck}. In the column of `Datasets', each line represents the traffic data of each point at three hours, in which the peak traffic value is highlighted in bold face. In the column of `Clustering', points with the numbers divided by the comma are in the same cluster. For example, `12,3' means first and second point are in the same cluster while the third point is an isolated cluster. In the last column, the best values obtained by the corresponding clustering scheme are highlighted in bold face. There are five different clustering choices shown in the second column. The values of $1-UC$ and $HC$ for those three clusters are presented in the third, fourth and fifth column of the table. If there is only two cluster, the values of $1-UC$ and $HC$ is shown in `NULL'. The clustering scheme with the largest values selected by $mean(1-U)$ or $meanM$ is also highlighted with bold font. 

After introducing the header of Table \ref{table:exampleHck}, the generated datasets are presented. In this table, there are six generated datasets. Among them, when doing the clustering on the first three datasets, there might be delay while there is no delay on the last three datasets. For example, when clustering all of the three points in the first dataset together, there are delays in three hours. However, no matter how to cluster the three points on the last three datasets, there is no dealy. In addition, the difference among the first three and the last three datasets is in the peak distribution of the traffic data and the peak traffic data among three hours is highlighted with bold font. 

It is clear from this table that $meanM$ cannot distinguish all choices on datasets where all points have the same peak hour. In addition, both $meanM$ and $mean(1-U)$ can get the best clustering on datasets where there is no delay in any clustering scheme. However, on datasets where there might be delay, $mean(1-U)$ gets best solution that makes more sense while $meanM$ tends to cluster all points together no matter whether there is delay. To conclude, the peak distribution in the existing problem formulation is unnecessary.

\section{SplitEA: An Evolutionary Approach to the Computing Resource Allocation Problem}
\label{sec:proposedEA}


As a meta-heuristic algorithm with capability of global search, evolutionary algorithms have been successfully applied to solve many resource allocation problems \cite{perveen2021dynamic} \cite{robinson2011market} \cite{lewis2009evolutionary} \cite{lewis2010resource} \cite{wang2010multi} \cite{lewis2008evolutionary} \cite{salcedo2008optimal} \cite{salcedo2008assignment}. Even though this, the existing evolutionary algorithms for solving resource allocation problems in the literature are not suitable for the problem in this paper, since the problem formulation in this paper is different from that in the literature and representation and evolutionary
operators of existing evolutionary algorithms are not directly applicable. More specifically, solution representation, genetic operators to generate offspring solutions and constraints handling mechanisms in existing evolutionary algorithms are not appropriate for the problem formulation in this paper. 
Except for the proposed EA components, the process of random cluster splitting in SplitEA is designed to produce an initial population for the problem of the next day, through randomly selecting a cluster from a solution in the optimized population of the previous day and randomly splitting it into two clusters. This section aims to answer the second research question: How to design an evolutionary algorithm tailored for solving the new formulation? Therefore, in this section, the specific process of the proposed SplitEA tailored for solving the new problem formulation are presented in detail.

\begin{algorithm}[ht!]
\KwIn{A set of $N$ points $R = (r_1,...,r_i,...,r_N)$ with their position coordinate $(r_i^1, r_i^2)$ (where $r_i^1$ and $r_i^2$ are the longitude and latitude, respectively); neighborhood threshold $\tau$; population size $popsize$; maximum number of generations $maxgen$; start date $startdate$ and end date $enddate$.}
\KwOut{The found best solutions $\textit{\textbf{x}}^*$ for days from $startdate + 1$ to $enddate + 1$.}
\nl Calculate the distance of any two points $dist_{i,j}$ to fill the distance matrix $distM = {dist_{i,j}}$\;
  \nl Initialize parameters: set the count of generation $gencount$ = 1; set the count of days $d$ = $startdate$\;
  \nl \textbf{InitialPop()}: randomly generate a population $\textbf{P}$ with a set of $popsize$ feasible solutions\;
  \nl \While{$d \leq enddate$}{
  \nl Input the traffic data of each point $\textit{\textbf{f\,}}_i^d = \big( f_0^d(r_i),..., f_h^d(r_i),...,f_{23}^d(r_i) \big)^T$ on the current day, where $f_h^d(r_i)$ is sum of data of $r_i$ from $h$-th to $(h+1)$-th hour at the current day\;
  \nl Utilize a forecasting model to predict the traffic data of each point on the next day ($(d+1)-$th) $\textit{\textbf{fp}}_i^ {d+1} = \big( f_0^{d+1}(r_i),..., f_h^{d+1}(r_i),...,f_{23}^{d+1}(r_i) \big)^T$ based on the current day's ($d-$th) traffic data $\textit{\textbf{f\,}}^{d}_i$ \;
  \nl Calculate the fitness value of each solution in  population $\textbf{P}$ based on the predicted traffic data $\textit{\textbf{fp}}_i^ {d+1}$ using the equation (\ref{eq:def of problem})\;
	\nl \While{$gencount \leq maxgen$}{
               \nl Generate an offspring population $\textbf{P}_{off}$ using the mutation operator based on the parent population $\textbf{P}$: \textbf{Mutation($\textbf{P}$)}\;
       \nl Calculate the fitness value of each solution in  population $\textbf{P}_{off}$ based on the predicted traffic data $\textit{\textbf{fp}}_i^ {d+1}$ using the equation (\ref{eq:def of problem})\;
        \nl Combine two populations $\textbf{P}$ and $\textbf{P}_{off}$ and select the top $popsize$ solutions as the $\textbf{P}$ for the next iteration\;
       \nl  $gencount$ = $gencount$ + 1\; 
      }
      \nl Select $\textit{\textbf{x}}_{d+1}$ with the smallest fitness value from $\textbf{P}$ as the solution to be adopted/deployed for day d+1\;
      \nl \textbf{RandomClusterSplitting($\textbf{P}$)}: conduct random cluster splitting on $\textbf{P}$ to produce a new population as the initial population $\textbf{P}$ for the next day\;
      \nl d = d + 1\;
      }
 
    \caption{{\bf Proposed SplitEA framework} \label{Alg:framework}}
\end{algorithm}

\subsection{Overview of the Proposed Evolutionary Algorithm SplitEA}

This section introduces the framework of the proposed evolutionary algorithm, which is exhibited in Algorithm \ref{Alg:framework}. Given a set of $N$ points with their position coordinate in the network,
the algorithm firstly calculates the distance of any two points to fill the adjacent matrix in line 1 of Algorithm \ref{Alg:framework}. Then, in line 2, the initialization process of parameters is conducted to set the initial values for the parameters used in SplitEA. The initial values of $gencount$ and $d$ are set as 1 and $startdate$, respectively. After that, the initialization process of the algorithm randomly generates a population $\textbf{P}$ with a set of feasible solutions in line 3. Then, allocate the computing resources to all points day to day from the start date $startdate$ to the end date $enddate$. Within the outer loop of the while from line 4 to line 15, the first step is to input the traffic data of all points on the current day in line 5. Then in line 6, leverage a prediction model to forecast the traffic data of all points on the next day. The input of the prediction model is the traffic data of all points on the current day. Note that, in reality, the traffic data of all points is only completely collected and therefore available at the end of each day. Next step in line 7 is to calculate the fitness value of all solutions in the initial population $\textbf{P}$ for this day on the predicted traffic data using the equation (\ref{eq:def of problem}).

\begin{algorithm}[ht!]
\KwIn{A set of $N$ points $R = (r_1,...,r_i,...,r_N)$ with their position coordinate $(r_i^1, r_i^2)$ (where $r_i^1$ and $r_i^2$ are the longitude and latitude, respectively); distance matrix $disM$; neighborhood threshold $\tau$;}
\KwOut{The initialized population $\textbf{P}$.}
 \nl \For{j:=1 to $popsize$}{
 \nl Set the cluster count $clucount$ as 0\;
	\nl \While{there is a point not clustered}{
         \nl Randomly select a point $r$ from the set of points $R$\;
         \nl Find all points from the rest points to get a set $CloseSet$, in which the distance of any point to $r$ is smaller than $\tau$, the size of the set $CloseSet$ is $N_c$\;
         \nl Randomly pick points from $CloseSet$ with the number of less than or equal to $N_c$ and group them with $r$ as a cluster\;
         \nl Set the value of those clustered points as: $x_k$=$clucount$, where $k$ is the index of those points grouped into the same cluster\;
         \nl $clucount = clucount + 1$\;
         }
       \nl  Set the $j-$th solution as $\textbf{P}_j = (x_1^j,...,x_N^j)$\;
         }
 \nl \textbf{Return} $\textbf{P}$.
    \caption{{\bf InitialPop(): procedures of the population initialization.} \label{Alg:initial}}
\end{algorithm}

After the initialization, the algorithm iterates until the $gencount$ reaches the pre-setting maximum number of generations within the loop of while from line 8 to line 12. At each generation, the proposed mutation operator is used to produce an offspring population $\textbf{P}_{off}$ with feasible solutions based on the parent population $\textbf{P}$ in line 9. Then in line 10, solutions in the offspring population are also evaluated using the fitness function in equation (\ref{eq:def of problem}). After that, the parent population $\textbf{P}$ and the offspring population $\textbf{P}_{off}$ are combined together in line 11. Then also in line 11, the top $popsize$ solutions with the smallest fitness values are selected from the combined population as the $\textbf{P}$ for the next generation. After the iteration in line 13, a solution with the smallest fitness value is selected from $\textbf{P}$ as the found best solution for the resource allocation problem at the $d+1$-th day to be deployed. Then, in line 14, conduct the process of random cluster splitting on $\textbf{P}$ to produce a new population as the initial population $\textbf{P}$ for the next day. This process is trying to improve the clustering structure diversity of the whole population such that better solutions can be found for the next day.



\subsection{Description of Each Step of the Algorithm}

This section describes the details of three procedures in the framework of the proposed evolutionary algorithm: the initialization process of the population InitialPop(), mutation operator to produce feasible offspring solutions Mutation($\textbf{P}$) and the process of random cluster splitting to produce the initial population for the next day RandomClusterSplitting($\textbf{P}$). These procedure are specifically designed tailored for solving the resource allocation problem. Considering that the constraint of this problem is very special, to make the algorithm concise, the constrain handling strategy is not developed. Alternatively, those three procedures are just designed to produce feasible solutions considering the constraint.

\paragraph{Population initialization} The detailed procedures of the population initialization are stated in Algorithm \ref{Alg:initial}, which aims to produce a population with $popsize$ feasible solutions. The basic idea is to randomly group a set of random number of neighboring points together to form a feasible solution. The specific steps for producing each feasible solution are as follows. Firstly, a point $r$ is randomly selected from the set of points in line 4. Then, find all neighboring points of $r$ to form a neighboring set of this point $CloseSet$ in line 5. All those neighboring points have the distance to $r$ smaller than $\tau$.  After that, in line 6, randomly pick several points with the number of no larger than $N_c$ to cluster them to point $r$, where $N_c$ is the size of $CloseSet$. Repeat these steps until all points in the set of all points are clustered. This initialization process aims to generate a population with the number of cluster smaller than the number of points and potentially increase the clustering structure diversity as much as possible, such that the optimal clustering structure could be found in later optimization process.

\begin{algorithm}[ht!]
\KwIn{The parent population $\textbf{P}$ distance matrix $disM$; neighborhood threshold $\tau$; the probability of preferentially clustering isolated points $prob$.}
\KwOut{The offspring population $\textbf{P}_{off}$.}
 \nl \For{j:=1 to $popsize$}{
 \nl Select the isolated points as the set $isoPoint$ from $\textbf{P}_j$, each of which solely forms a cluster\;
 \nl Random generate a number $randNum$ between 0 and 1: $randNum = rand(0,1)$ \;
 \nl \eIf{$randNum < prob$ \textit{and} $isoPoint$ is not null}
                {
                      \nl  Randomly select an isolated point $x$ with the index $k$ from the set $isoPoint$\;
                }
                {
                  \nl  Randomly select a point from all points\;
                }
                
             \nl Find all adjacent clusters of $x$ as $mutClusters$, in which all points have the distance to point $x$ smaller than or equal to $\tau$\;  
              
 \nl \eIf{$mutClusters$ is not NULL}
                {
                \nl Group $x$ into a random cluster $randCluster$ in $mutClusters$ and set the value of $x_k$ as the cluster number of $randCluster$: $x_k = randCluster$ \;
                }
                {
                  \nl  Randomly select an adjacent cluster $C$ and put those points whose distance to $x$ is smaller than or equal to $\tau$ in the set $C_{close}$, of which the size is $Num$\;
                  \nl Randomly pick a random number of points between 1 and $Num$ from $C_{close}$, and group them and $x$ into a new cluster\;
                  \nl Set the value of those points in  the new cluster as  $max(x_i)+1, (x = 1,...,N)$ (It is impossible that some existing clusters are eliminated since the size of cluster $C$ is larger than $Num$. The reason is $mutClusters$ is NULL.)\;
                }
       \nl  Get the $j-$th mutated solution $\textbf{P}_j = (x_1^j,...,x_N^j)$\;
         }
 \nl \textbf{Return} $\textbf{P}_{off}$.
    \caption{{\bf Mutation($\textbf{P}$): procedures of the mutation operator.} \label{Alg:mutation}}
\end{algorithm}

\paragraph{Mutation operator}
It has been attempted to design a crossover operator to generate feasible solutions from two parents. However, it is very strict and difficult to generate feasible solutions as the point $x_i$ to be swapped must have the distance smaller than or equal to any points in the cluster whose number is going to be assigned to $x_i$. Therefore, only the mutation operator is utilized. As the only genetic operator in the proposed evolutionary algorithm, the proposed mutation operator combines the role of both crossover and mutation operator in normal evolutionary algorithms. How the designed mutation operator achieves this goal is presented in Algorithm \ref{Alg:mutation}. Given the parent population $\textbf{P}$, the mutation operator produces an offspring population $\textbf{P}_{off}$ with $popsize$ feasible solutions. For each solution of $\textbf{P}$, the operator firstly select the isolated point as the set of $isoPoint$ in line 2, in which each point is formed as one isolated cluster. Here, isolated points are those that are not grouped into any clusters with other points and each of them form a single cluster. Then, in line 3, randomly generate a random number $randNum$ between 0 and 1. If $randNum$ is smaller than the pre-setting probability $prob$ and the set $isoPoint$ is not empty, randomly select an isolated point $x$ with the index $k$ from the set $isoPoint$ in line 5; else, randomly select a point from all points in line 6. Here, $prob$ controls the weight of decreasing the number of clusters and increasing the diversity of the clustering scheme. For example, if $prob$ is large, more isolated points could be given more priority to be grouped to its adjacent clusters.

Afterwards, the mutation process is conducted on the  selected point $x$, trying to set $x_k$ as the cluster number to produce a feasible solution. Firstly, find all adjacent clusters of $x$ as $mutClusters$ in line 7, in which all points have the distance to point $x$ smaller than or equal to $\tau$. This is to ensure whether $x$ can be directly grouped to the existing cluster. If the set $mutClusters$ is not null, just directly group $x$ into one of the randomly picked cluster in the $mutClusters$ to decrease the number of clusters in line 9; else, in line 10 randomly select an adjacent cluster $C$ of $x$ and put those points whose distance to $x$ is smaller than or equal to $\tau$ in the set $C_{close}$ with the number of $Num$. After that, randomly pick a random number of points between 1 and $Num$ from $C_{close}$ group them and $x$ into a new cluster in line 11. Following this way, feasible solutions in the offspring population $\textbf{P}_{off}$ can be produced. In the second case, a new cluster is produced, which results in one more cluster while increase the diversity of the clustering. Therefore, solutions with more resource utilization rate and less delay might be searched.

\paragraph{Random Cluster Splitting}

Given that problems at different days have the same position information of all points, solutions found for the previous day might be useful for the problem with the next day. Bearing this in mind, we propose to conduct a random cluster splitting process on the found solutions of the problem at the previous day, so as to generate a novel population as the initial population for the optimization of the problem at the next day. The main idea is to split a randomly selected cluster into two clusters, in which each cluster has the random number of points. The reason why not just copying the individuals from the previous population is that copied population has very few diversity regarding the number of cluster, which may prevent the search of solutions with better fitness value and few BBU underutilization and delay. The specific procedures of the random cluster splitting is described in Algorithm \ref{Alg:randVariation}.

For each solution in the population $\textbf{P}$, conduct the following random cluster splitting on it to increase the structure diversity of the existing clustering scheme, such that each solution has more chance to reach better fitness in later optimization. Firstly, in line 2, randomly pick a cluster $C$ with more than one point as the cluster to be split, since there is no need to split with just one point. Then, calculate the number of points in the cluster C: $NR$ in line 3. Next, in line 4, randomly generate a number between 1 and $\lfloor NR/2 \rfloor$: $Nsplit = rand(1, \lfloor NR/2 \rfloor)$. This also enables the diversity of the clustering structure. After that, in line 5, split the cluster C into two clusters with size of $Nsplit$ and $NR- Nsplit$, respectively. Lastly, set the value of those points in the cluster with the size of $NR- Nsplit$ as $max(x_i)+1, (x = 1,...,N)$. Through this random cluster splitting process, the produced solutions can decrease the number of cluster by 1 and therefore maintain most clustering structure. At the same time, the diversity of the clustering structure can be increased through splitting the cluster, which might generate better solutions for the problem at the next day. 

\begin{algorithm}[ht!]
\KwIn{The population $\textbf{P}$ of the problem at the previous day; distance matrix $disM$; neighborhood threshold $\tau$;}
\KwOut{The initialized population $\textbf{P}$ for the problem at the next day.}
 \nl \For{j:=1 to $popsize$}{
 \nl Randomly select a cluster $C$ with more than one point\;
 \nl Calculate the number of points in the cluster C: $NR$\;
 \nl Randomly generate a number: $Nsplit = rand(1, \lfloor NR/2 \rfloor) $\;
 \nl Split the cluster C into two clusters with size of $Nsplit$ and $NR- Nsplit$, respectively\;
 \nl Set the value of those points in the cluster with the size of $NR- Nsplit$ as $max(x_i)+1, (i = 1,...,N)$\;
       \nl  Set the $j-$th solution as $\textbf{P}_j = (x_1^j,...,x_N^j)$\;
         }
 \nl \textbf{Return} $\textbf{P}$.
    \caption{{\bf RandomClusterSplitting($\textbf{P}$): procedures of the random cluster splitting.} \label{Alg:randVariation}}
\end{algorithm}

\subsection{Time Complexity of the Proposed Evolutionary Algorithm}
This section analyzes the time complexity of the proposed knowledge transfer-based evolutionary algorithms. The processes that consume main time complexity will be analyzed. Step 1 in Algorithm \ref{Alg:framework} takes $O(N^2)$ computations, where $N$ is the number of points. The while loop from Step 3 to Step 8 in Algorithm \ref{Alg:initial} takes less than $O(N)$. Therefore, the population initialization consumes maximum $O(N*popsize)$ computations. At each generation of the evolutionary algorithm, mutation process consumes the most time complexity. Step 7 of the Algorithm \ref{Alg:mutation} consumes $O(N)$ computations while all other steps in the loop from Step 2 to Step 12 in Algorithm \ref{Alg:mutation} consumes less than $O(N)$ computations. Therefore, the mutation process consumes $O(N*popsize)$. Steps 2 to 7 in the Algorithm \ref{Alg:randVariation} consumes $O(NR-Nsplit)$, as Step 6 consists a for loop taking $O(NR-Nsplit)$ computations. Therefore, the process of random cluster splitting takes $O(popsize)$ computations. To conclude, the proposed evolutionary algorithm except for the prediction model consumes $O(days*maxgen*N*popsize)$ computations, where $days$ is the number of days in the dataset and $popsize$ is the population size.

\section{Experimental Studies}

This section introduces the used datasets, compared algorithms, parameter settings for them and performance metrics in the experimental studies of this paper, so as to verify the effectiveness of the proposed evolutionary algorithm. Specifically, Section \ref{sec:ComSplitEAGreedy} tries to answer the first sub-research question of the second research question: 
Does proposed EA outperform a greedy algorithm? Under what conditions. Then, the second sub-research question that what is the influence of different algorithm's design choices on its performance is answered in Section \ref{sec:ComEAs}.


\subsection{Experimental Setup}

\subsubsection{Datasets Description}
\label{sec:datasets}

There are two sets of datasets to be used, one of which is the real-world datasets found online while another is the artificial datasets. The real-world datasets are used to verify whether the proposed approach is able to solve the real-world problems better than the greedy algorithm. However, the available real-world datasets have limited types of location and traffic dataset. In order to verify the effectiveness of the proposed algorithm on datasets with more properties, several artificial datasets with the combination of different location and different traffic dataset are generated.

\paragraph{Real-world Datasets}
There are four real datasets found in the literature \cite{barlacchi2015multi} \cite{du2019inter} \cite{chen2015analyzing} \cite{rodoshi2020deep}. However, considering that Archive dataset does not have the location information, three location datasets are generated to complement it. Therefore, there are six real-world datasets in total. The datasets are explained below:
\begin{itemize}
\item The dataset of Milan includes two months of network traffic data from 11/01/2013 to 12/31/2013 from the Telecom Italia Big Data Challenge dataset \cite{barlacchi2015multi}. The city of Milan is partitioned into $100 \times 100$ grids with grid size of about $235 \times 235$ square meters. In each grid,
the traffic volume is recorded on an hourly basis. We compile a base station dataset from CellMapper.net, which consists of the locations and coverage areas of active base stations observed in the two months. Based on the location and coverage of each base station, we find the corresponding covered grids and calculate their traffic volume. Finally, we normalize the traffic volumes of each base station to the [0;1] range for the convenience of analytics.
\item The Songliao Basin dataset \cite{du2019inter} contains movements of near 3-million anonymized cellular phone users among 167 divisions (henceforth locations), covering 4 geographically adjacent areas (Changchun City, Dehui City, Yushu City, and Nong’an County) for a one-week period starting on August 7, 2017. Tis total geographic area, located in the southeast Songliao Basin in the center of the Northeast China Plain, Northeast China, covers more than 20 square kilometers. It has two files, one of which is `Mobility.txt' describing the hourly-mobility network for the entire week. In this file, each row represents the total number of hourly movements by people from locations $i$ to $j$ in the corresponding day.  Another file 'GPS.txt' includes the latitude and longitude information for each location in the mobility network. The pre-processing on the 'Mobility.txt' is to calculate the data of each point through adding all weights of coming to this point and 
originating from this point. After that, if there is one or more hours when some points do not have the data, delete those points from the 'GPS.txt' and delete those rows with the data for those points. Finally, normalize the data of each point to the $[0,1]$ range.
\item C2TM \cite{chen2015analyzing}: This dataset consists of individuals' activities during a continuous week (August 20 to
26, 2012), with accurate timestamps and location information indicated by the longitude and latitude of connected points. From geographic view, the monitored
link covers a cellular area of around 50km×60km. In order to specify principle spatio-temporal properties of cellular traffic a subarea, around 28km $*$ 35km including more than 85\% of total population in both city center and suburbs, is selected for our analysis. Specifically, the selected part consists of 13K
BTSs that serve more than 452K users, totally generating 379 millions of HTTP records in the measurement week.
\item Archive \cite{rodoshi2020deep}: this dataset has 57 points and the data is collected in approximately 1 year x 24 hours x 57 points. However, there are some hours when not all points have the data. Therefore, points without traffic data are deleted. There are 54 points left with the date from 12/02/2018 to 20/07/2018. In addition, this dataset does not have the location information. Therefore, three location datasets are generated. The details of the generation is explained as follows:
\begin{itemize}
\item 54 points are randomly selected from the location dataset of Milan;
\item 54 points are randomly selected from the location dataset of Songliao;
\item The third location dataset is randomly generated within a range with 54 points.
\end{itemize}
Those three datasets are named as Archive-Milan, Archive-Songliao and Archive-Random, respectively.
\end{itemize}

\paragraph{Artificial Datasets}

This paragraph describes the generated artificial datasets and how they are combined with different types of generated location dataset and generated traffic dataset. There are 8 generated artificial datasets with seven days, which includes 1a, 2a, 3a 100/158, 3a 120/158, 1c-Milan, 1c-Songliao, 2b-Np=10 (Nt=174) and 2b-Np=5(Nt=185), where $Np$ is the maximal number of points in each group for the location dataset, $Nt$ is the total number of points in the solution. Note that in dataset 2a, the maximal number of points ($Np$) in each group for the location dataset is set as 5. Among those datasets, `1', `2' and `3' means the first, second and third location dataset, respectively; `a', `b' and `c' means the first, second and third traffic dataset, respectively.  Those datasets have been released online \footnote{The website will be written once the datasets are put in order online}.

\begin{itemize}
\item There are three location datasets with different types of location information:
\begin{enumerate}
\item All points are totally randomly generated within a range;
\item High cohesion and low coupling w.r.t distance of points. It means that those points that are close to each other are in the same group. More specific, the distance of any two points in the same group is smaller than $\tau$ and the distance of any two points in different groups is larger than $\tau$, the maximum number of generated points in each cluster is Np.
\item Many points are gathered together while others are scattered away from those points. The distance of any two gathered points is much smaller than that of gathered point and that scattered away. Ng and Nt are the number of gathered and total points, respectively.
\end{enumerate}
\item There are three traffic datasets with different traffic pattern:
\begin{enumerate}
\item Totally randomly generated from (0,1) for each point at 24 hours of each day;
\item Generate the traffic data in a way that the optimal value of the objective function is known for the second case of the location dataset. More specifically, for each cluster in which all points have the distance close to each other smaller than $\tau$, just split it into several sub-clusters and then generate the traffic data such that the total traffic data in each sub-cluster is equal to 1 at each hour of a day.
\item Follow the pattern of existing real dataset Milan and Songliao. Those patterns are extracted from the traffic dataset of Milan and Songliao datasets. For the traffic pattern of Milan dataset, the traffic data for all points firstly decreases at the first five or six hours of each day and then increases until noon. Then, it remains stable at five or six hours and lastly it decreases. As for the traffic pattern of Songliao dataset, it firstly increases until eight or nine of each day and then remains stable for ten or eleven hour and lastly decreases. The traffic dataset based on Milan and Songliao datasets are generated following the traffic pattern of them, just as described before this sentence.
\end{enumerate}
\end{itemize}
The codes on specifically generating those datasets will be released on.
\subsubsection{Model Specification}

In the experiments on real-world datasets with the prediction, we use the LSTM model to do the prediction, which is used in Step 6 of Algorithm \ref{Alg:framework}. The reason why we use LSTM is that it can capture the temporal dependency and spatial correlation among base station traffic
patterns \cite{chen2018deep} and it has been proved to achieve better prediction results than ARIMA and WANN in \cite{chen2018deep}. The specification of the model and the model parameters are the same as those used in \cite{chen2018deep}. The model has two stacked LSTM layers. The encoder layer $L1$ contains $N_{en}$ encoder memory units, which accepts a traffic snapshot of shape [24;182] as input, and outputs an encoded sequence for the decoder. The decoder contains $N_{de}$ decoder memory units, which accepts the encoded sequence as input and outputs the forecast of the traffic snapshot.

Considering that only the Milan and Archive datasets have many days to do the prediction, compared algorithms on those two days will use the LSTM to do the prediction. The division of the training set and testing set is also the same as that in \cite{chen2018deep}, i.e. the first 70\% days of the dataset is regarded as the training set with the remaining days as the testing set. For other datasets except Milan and Archive which do not have enough days, no splitting process is conducted on them, which are not used to do the prediction.


\subsubsection{Compared Algorithms}
\label{sec:comAlgo}
There are four compared algorithms in the following experiments.
\begin{enumerate}
\item The greedy algorithm (GreedyAlg): modified from the greedy algorithm in the existing work \cite{chen2018deep} through applying the fitness function of the proposed problem formulation and setting the termination criteria as the pre-set maximum number of evaluations. In the greedy algorithm of the existing work \cite{chen2018deep}, it calculates the fitness value of clusters formed by a selected point to its adjacent clusters, of which more details can be found in \cite{chen2018deep}. However, in the greedy algorithm here, it calculates the fitness function of the cluster schemes (solutions) formed by a selected point to its adjacent clusters as well as other clusters. The greedy algorithm here is regarded as a baseline algorithm to verify the effectiveness of the proposed evolutionary algorithms. The aim of setting the maximum number of evaluations  as the termination criterion is to make a fair comparison to the proposed evolutionary approach through setting the evaluation times the same for all compared algorithms.
\item RandEA: replace step 14 of Algorithm \ref{Alg:framework} with Algorithm \ref{Alg:initial}: it randomly generated an initial population whenever solving the problem of the next day; other procedures remain the same. The reason why to introduce this algorithm is to verify whether the random cluster splitting is better than optimization from scratch.
\item CopyEA: replace step 14 of Algorithm \ref{Alg:framework} with the following process: just copy $\textbf{P}$ as the initial population $\textbf{P}$ for the next day. CopyEA is regarded as the compared algorithm to prove whether the random cluster splitting process is better than simply copying all previous solutions.
\item SplitEA: our proposed evolutionary algorithm.
\end{enumerate}
In order to verify the performance of our proposed SplitEA over the greedy algorithm, we compare SplitEA against the greedy algorithm on six real-world datasets and eight artificial datasets. In addition, we analyze the impact of different parameter settings on the performance of SplitEA and the greedy algorithm on Milan dataset. The comparison results are shown in Section \ref{sec:ComSplitEAGreedy}. Similarly, the performance of different algorithm design choices and the influence of parameter settings on them are also analyzed. The comparison results are shown in Section \ref{sec:ComEAs}.

\subsubsection{Parameter Settings}
In this section, we describe the used specific parameter settings, like the mutation probability and the parameters of the problem.
\begin{itemize}
\item $prob$=0.5 in the improved mutation, which makes the probability of clustering isolated points into a cluster and that of searching for different clustering structure equal.
\item Neighborhood threshold $\tau$ = $3d$, where $d$ is the average distance of all point to their closest point, which is selected based on the experience.
\item Evaluation time in GreedyAlg = 1500.
\item $w$ = 0.01 in the problem formulation; $popsize$ = 10 and $generation$ = 150 for all compared EAs including SplitEA, RandEA and CopyEA, which are set to make the evaluation times = 1500, equal to that of the greedy algorithm to enable fair comparison. The evaluation time, generation and population size are selected based on the experience.
\item Mutation probability: 1, as there is only one operator to evolve the population;
\item Each algorithm is given 30 independent runs. Friedman and Nemenyi statistical tests \cite{demvsar2006statistical} with the significance level 0.05 are used to indicate the statistical significance of all compared algorithms. The metric value obtained by a given algorithm on one dataset is regarded as an observation to compose that algorithm’s group for the test, following Demsar’s guidelines \cite{demvsar2006statistical}. Therefore, there are 30 observations in each group.
\end{itemize}
The impact of these parameters will also be analyzed in Section \ref{sec:ComSplitEAGreedy} and Section \ref{sec:ComEAs}.
\subsubsection{Performance Metrics}
\label{Metrics}


Metrics to evaluate the performance of found solutions by the greedy algorithms and the evolutionary algorithms are shown as follows. They are set according to the optimization objectives of the problem, so that we can investigate how well the algorithm is able to optimize these objectives.
\begin{enumerate}
\item The number of clusters: K;
\item The average difference between all points’ data and 1 for all clusters: $U$, which is defined as follows:
\begin{equation}
U = \frac{1}{K}\sum\limits_{k=1}^{K}U(C_k)
\end{equation}
\item The average difference of all points data and 1 when there is delay: Udelay, which is described as follows:
\begin{equation}
Udelay = \frac{1}{K}\frac{1}{24}\sum\limits_{k=1}^{K}\sum_{}^{}\big(f_h(C_k)-1\big),  ~when ~f_h(C_k)>1
\end{equation}
\item The average difference of all points data and 1 when there is NO delay: Uunder1, which is described as follows:
\begin{equation}
Uunder1 = \frac{1}{K}\frac{1}{24}\sum\limits_{k=1}^{K}\sum_{}^{}\big(1-f_h(C_k)\big), ~when ~f_h(C_k)\leq1
\end{equation}
\end{enumerate}
Note that all metrics are required to be minimized. Minimizing Uunder1 means maximizing the utilization rate.


%
%
%
%
%
%
%

\subsection{Experimental Results}

This section firstly presents the comparing results between the proposed SplitEA and the greedy algorithm on the real-world datasets, to show the superiority of our proposal. Then, the comparison results of SplitEA and the greedy algorithm on artificial dataset are shown to verify the performance of them under different scenarios, like different distribution of points and different traffic pattern. Later on, three EAs (RandEA, CopyEA and SplitEA) are compared on both real-world and artificial datasets to verify the benefit of the process of random cluster splitting in SplitEA.

\subsubsection{Comparison Results of SplitEA and the Greedy Algorithm}
\label{sec:ComSplitEAGreedy}

\begin{table}[ht!]
\begin{center}
\caption{Comparison results of SplitEA and the greedy algorithm on real-world datasets.}
\label{tab:GreedySplitReal}
\begin{tabular}{|c|cc|cc|cc|}
\hline
Datasets & \multicolumn{2}{|c|}{Milan} &\multicolumn{2}{|c|}{Songliao} & \multicolumn{2}{c|}{C2TM-sub1} \\
\hline
Algorithms & GreedyAlg &	SplitEA &	GreedyAlg &	SplitEA & GreedyAlg &	SplitEA\\
\hline
K	&	81.3907 	&	\textcolor{red}{52.3019} 	&	104.7381 	&	\textcolor{red}{54.3619} 	&	59.6792 	&	\textcolor{red}{51.7708} 	\\
U	&	0.8388 	&	\textcolor{red}{0.7644} 	&	0.8715 	&	\textcolor{red}{0.8042} 	&	0.9961 	&	\textcolor{red}{0.9956} 	\\
Udelay	&	\textcolor{red}{0.0003} 	&	0.0076 	&	\textcolor{red}{0.0001} 	&	0.0234 	&	1.25E-07 	&	2.97E-07 	\\
Uunder1	&	0.8385 	&	\textcolor{red}{0.7568} 	&	0.8714 	&	\textcolor{red}{0.7760} 	&	0.9961 	&	\textcolor{red}{0.9956} 	\\
\hline
f	&	1.6527 	&	\textcolor{red}{1.2874} 	&	1.9189 	&	\textcolor{red}{1.3478} 	&	1.5929 	&	\textcolor{red}{1.5133} 	\\
\hline
\hline
Datasets & \multicolumn{2}{|c|}{Archive-Milan} &\multicolumn{2}{|c|}{Archive-Songliao} & \multicolumn{2}{c|}{Archive-Random} \\
\hline
Algorithms & GreedyAlg &	SplitEA &	GreedyAlg &	SplitEA & GreedyAlg &	SplitEA\\
\hline
K	&	15.3396 	&	\textcolor{red}{13.0352}	&	15.6082 	&	\textcolor{red}{13.2629} 	&	15.2642 	&	\textcolor{red}{12.0069} 	\\
U	&	0.7839 	&	\textcolor{red}{0.7536} 	&	0.7880 	&	\textcolor{red}{0.7538} 	&	0.7823 	&	\textcolor{red}{0.7260} 	\\
Udelay	&	\textcolor{red}{0.0001} 	&	0.0039 	&	\textcolor{red}{0.0002} 	&	0.0022 	&	\textcolor{red}{0.0001} 	&	0.0009 	\\
Uunder1	&	0.7837 	&	\textcolor{red}{0.7497} 	&	0.7878 	&	\textcolor{red}{0.7517} 	&	0.7822 	&	\textcolor{red}{0.7250} 	\\
\hline
f	&	0.9372 	&	\textcolor{red}{0.8840} 	&	0.9440 	&	\textcolor{red}{0.8865} 	&	0.9349 	&	\textcolor{red}{0.8460} 	\\
\hline
\end{tabular}
\vspace{-0.48cm}
\end{center}
\begin{tablenotes}
      \footnotesize
      \item There are 30 independent runs. The values in this table are the mean value of the metrics under 30 runs. Friedman and Nemenyi statistical tests \cite{demvsar2006statistical} with the significance level 0.05 are used to indicate the statistical significance between compared algorithms. The metric value obtained by a given algorithm on one dataset is regarded as an observation to compose that algorithm’s group for the test, following Demsar’s guidelines \cite{demvsar2006statistical}. Therefore, there are 30 observations in each group for each metric on each dataset. The significantly better values obtained by the algorithm are highlighted in red color.
    \end{tablenotes}
\end{table}

This section tries to answer the first sub-research question of the second question: Does proposed EA outperform a greedy algorithm? Under what conditions?
Specifically, this section presents the comparison results of SplitEA and the greedy algorithm on several real-world and artificial datasets, to show the superiority of our proposal over the greedy algorithm. Firstly, SplitEA and the greedy algorithm are tested on the real-world datasets by using the real traffic data in the fitness function instead of the predictions of such traffic data. This has the aim of evaluating the optimization mechanisms themselves in the proposed SplitEA.
Then, two algorithms are tested on two real-world traffic datasets (Milan and Archive) with predictions to see whether the prediction error affects the performance of SplitEA through using the predicted traffic data in the fitness function. 
Lastly, two compared algorithms are tested on artificial datasets with different scenarios, to further verify the effectiveness of the optimization mechanisms themselves in the proposed SplitEA on more datasets beyond the real-world datasets. Note for those experiments without using the prediction model, the Step 6 in Algorithm \ref{Alg:framework} is deleted and in Step 7, the fitness value of solutions is calculated on the real traffic data.

\paragraph{Comparison results of SplitEA and greedy algorithm on real datasets}The comparison results of the greedy algorithm and SplitEA on the real dataset are shown in Table \ref{tab:GreedySplitReal}. It is clear that SplitEA gets significant better fitness values on all datasets, which shows that the SplitEA is able to get better solutions as expected. In addition, SplitEA achieves significant better values of all 4 metrics on all datasets except for the metric $Udelay$. The reason why the compared two algorithms get equal $Udelay$ on the dataset $C2TM-sub$ has been analyzed that the traffic data of most points at most hours of each day is too small to cause delay. This can be reflected by the results of the computing resource utilization rate ($Uunder1$), as the traffic data is too small to make the BBU rather underutilization. For all other datasets, SplitEA gest worse significant $Udelay$. It is intuitive to get the reason that SplitEA tends to cluster more points together due to the less required number of clusters for fixed number of points and this would inevitably increase the delay in the network.


\begin{figure}[!t]
  \centering
  \subfigure[The base station located in a residential
district of Milan dataset from 12/14/2013 to 12/31/2013.]{
    \includegraphics[width=.487\textwidth]{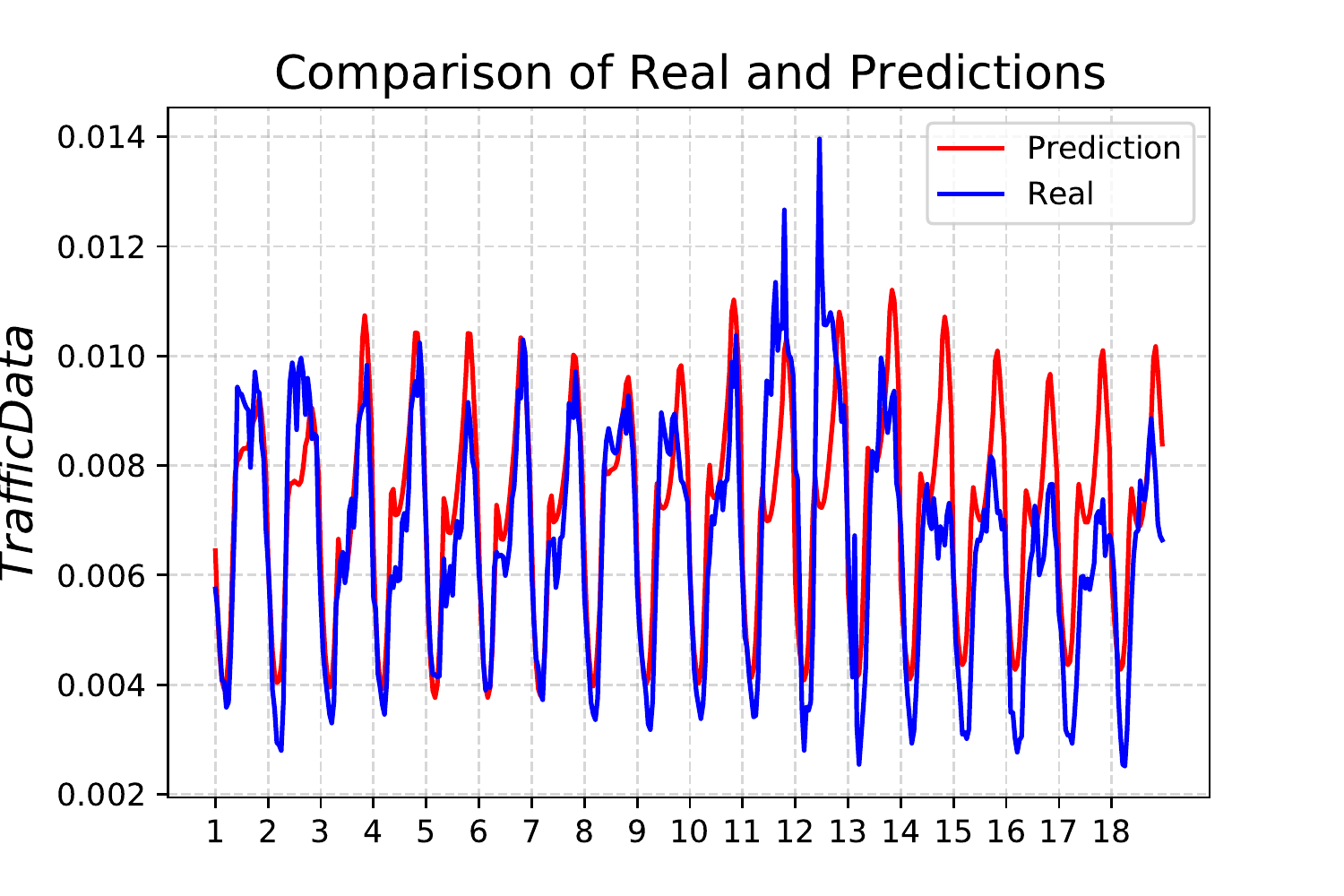}}
  \subfigure[The base station located in a business
district of Milan dataset from 12/14/2013 to 12/31/2013.]{
    \includegraphics[width=.487\textwidth]{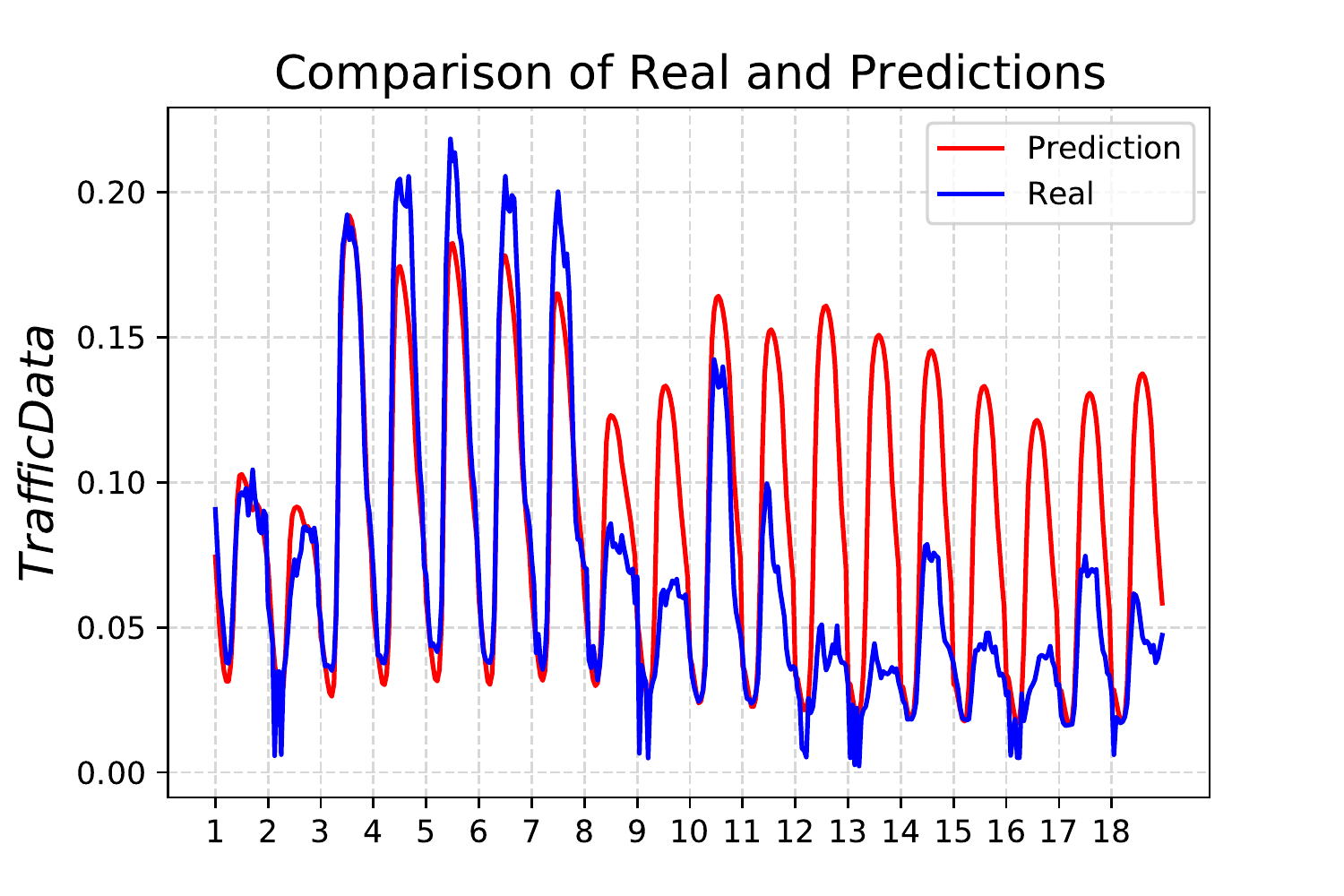}}
    \vspace{-0.4cm}
    \subfigure[A random picked base station of Archive dataset from 1/6/2018 to 14/6/2018.]{
    \includegraphics[width=.487\textwidth]{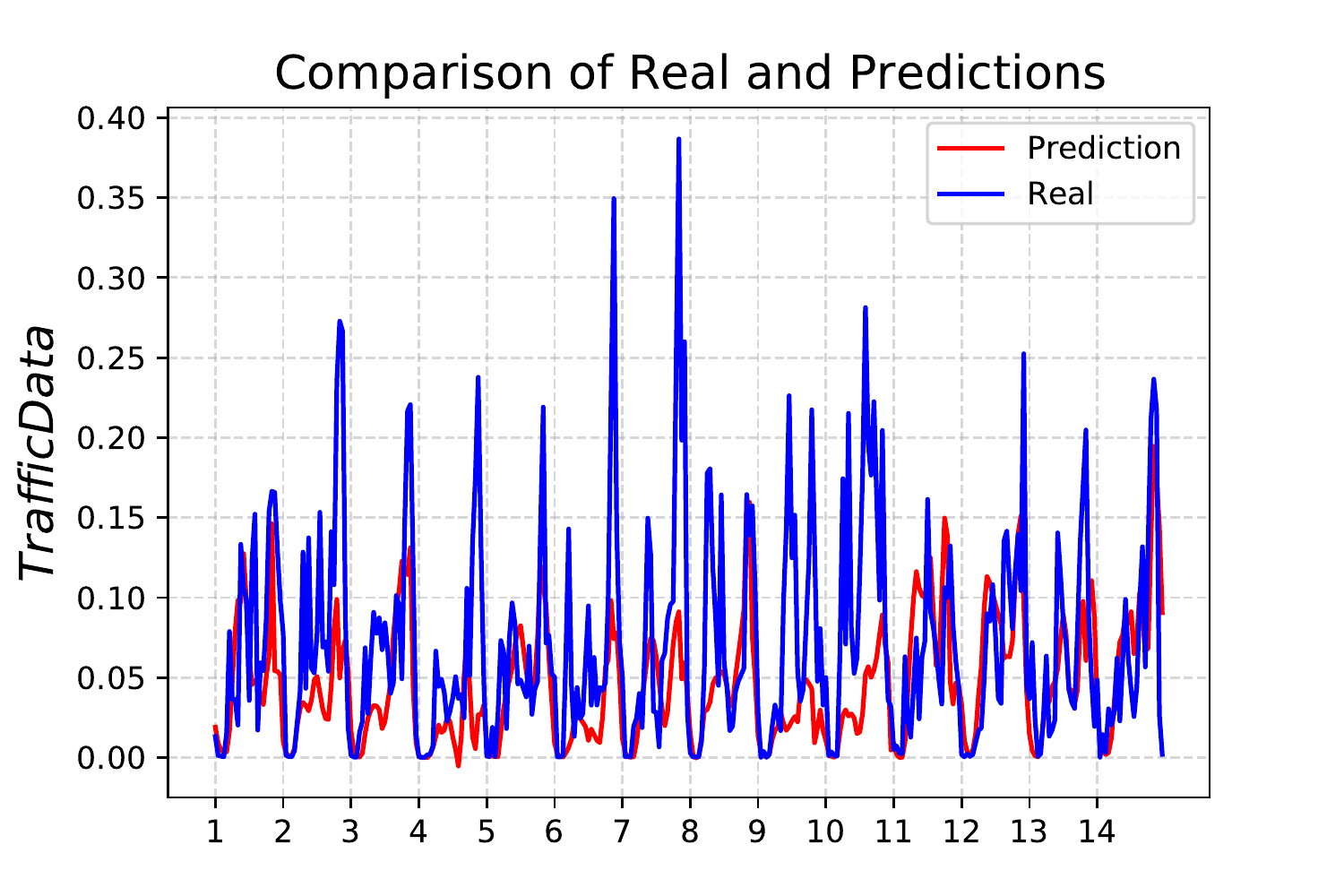}}
  \subfigure[A random picked base station of Archive dataset from 1/6/2018 to 14/6/2018.]{
    \includegraphics[width=.487\textwidth]{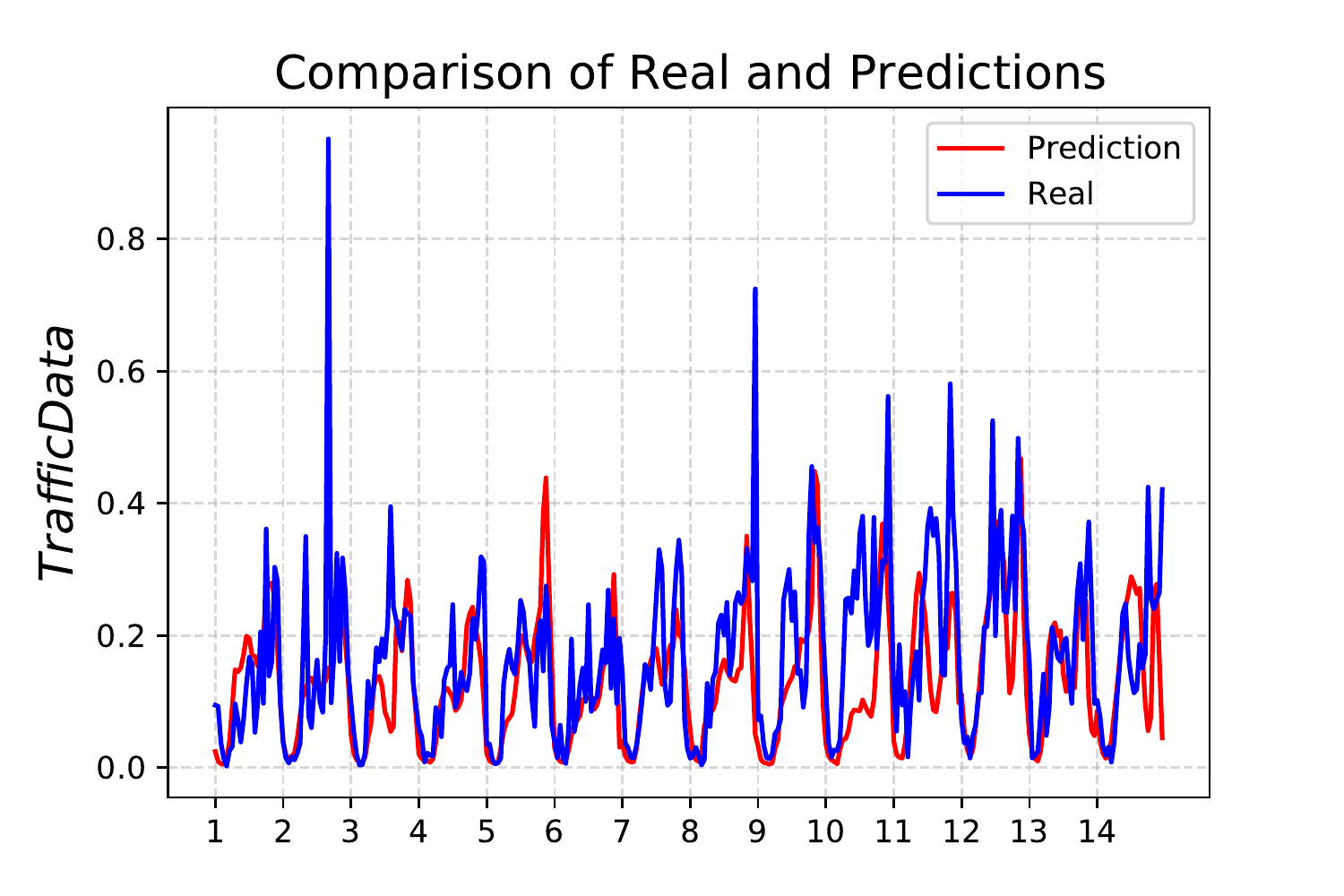}}
  \caption{RRH traffic prediction results by LSTM for two base stations of Milan and Archive datasets.}
  \label{fig:LSTMprediction} 
\end{figure}

\paragraph{Prediction results of LSTM on Milan and Archive Datasets} In order to check how much the optimization process of the proposed SplitEA and the greedy algorithm is affected by the prediction errors, this paragraph presents the prediction errors of the LSTM model on two real-world datasets Milan and Archive. The RRH traffic prediction results by LSTM for two randomly picked base station from Milan and Archive datasets are shown in Figure \ref{fig:LSTMprediction}. For Milan dataset, a residential and a business areas are picked with the results shown in Figure \ref{fig:LSTMprediction} (a) and (b), respectively. For Archive dataset, two randomly selected picked base stations with the results shown in Figure \ref{fig:LSTMprediction} (c) and (d), respectively. It is clear from \ref{fig:LSTMprediction} (b) that at the week of Christmas, the prediction errors on the business ares is much larger than that of other days.

\begin{table}[ht]
\scriptsize
\begin{center}
\caption{Comparison results of SplitEA and the greedy algorithm on real-world datasets with the prediction.}
\label{tab:GreedySplitPred}
\begin{tabular}{|c|cc|cc|cc|cc|}
\hline
Datasets & \multicolumn{2}{|c|}{Milan} &\multicolumn{2}{|c|}{Archive-Milan} & \multicolumn{2}{c|}{Archive-Songliao} & \multicolumn{2}{c|}{Archive-Random} \\
\hline
Algorithms & GreedyAlg &	SplitEA &	GreedyAlg &	SplitEA & GreedyAlg &	SplitEA & GreedyAlg &	SplitEA\\
\hline
K	&	81.4481 	&	\textcolor{red}{52.4222} 	&	14.9969 	&	\textcolor{red}{13.0181} 	&	14.4126 	&	\textcolor{red}{12.8106} 	&	15.0478 	&	\textcolor{red}{12.0019}	\\
U	&	0.8389 	&	\textcolor{red}{0.7673} 	&	0.7811 	&	\textcolor{red}{0.7540} 	&	0.7841 	&	\textcolor{red}{0.7553} 	&	0.7814 	&	\textcolor{red}{0.7267} 	\\
Udelay	&	\textcolor{red}{0.0003} 	&	0.0089 	&	\textcolor{red}{0.0013} 	&	0.0043 	&	0.0074 	&	\textcolor{red}{0.0072} 	&	\textcolor{red}{0.0012} 	&	0.0013 	\\
Uunder1	&	0.8386 	&	\textcolor{red}{0.7585} 	&	0.7799 	&	\textcolor{red}{0.7497} 	&	0.7767 	&	\textcolor{red}{0.7480} 	&	0.7802 	&	\textcolor{red}{0.7253} 	\\
\hline
f	&	1.6534 	&	\textcolor{red}{1.2915} 	&	0.9311 	&	\textcolor{red}{0.8842} 	&	0.9282 	&	\textcolor{red}{0.8834} 	&	0.9319 	&	\textcolor{red}{0.8467} 	\\
\hline
\end{tabular}
\vspace{-0.48cm}
\end{center}
\begin{tablenotes}
      \footnotesize
      \item There are 30 independent runs. The values in this table are the mean value of the metrics under 30 runs. Friedman and Nemenyi statistical tests \cite{demvsar2006statistical} with the significance level 0.05 are used to indicate the statistical significance between compared algorithms. The metric value obtained by a given algorithm on one dataset is regarded as an observation to compose that algorithm’s group for the test, following Demsar’s guidelines \cite{demvsar2006statistical}. Therefore, there are 30 observations in each group for each metric on each dataset. The significantly better values obtained by the algorithm are highlighted in red color.
    \end{tablenotes}
\end{table}

\paragraph{Comparison results of SplitEA and greedy algorithm on real-world datasets with the prediction}The comparison results of the greedy algorithm and SplitEA on the real-world datasets with predicted traffic data are shown in Table \ref{tab:GreedySplitPred}. Note that only Milan and Archive datasets have enough days to do the prediction. Therefore, there are only four datasets used to test the ability of compared greedy algorithm and SplitEA on the predicted traffic datasets. It is clear from Table \ref{tab:GreedySplitPred} that SplitEA gets significant better fitness value on all predicted datasets. In addition, SplitEA gets significant better metrics value regarding all four metrics on $Archive-Songliao$. On other datasets, SplitEA gets significant better metrics regarding all metrics except for $Udelay$. It is intuitive to get the reason that SplitEA tends to cluster more points together due to the less required number of clusters for fixed number of points and this would inevitably increase the delay in the network.


\begin{table}[ht]
\scriptsize
\begin{center}
\caption{Comparison results of SplitEA and the greedy algorithm on artificial datasets.}
\label{tab:GreedySplitArtif}
\begin{tabular}{|c|cc|cc|cc|cc|}
\hline
Datasets & \multicolumn{2}{|c|}{Dataset 1a} &\multicolumn{2}{|c|}{Dataset 2a} & \multicolumn{2}{c|}{Dataset 3a 100/158} & \multicolumn{2}{c|}{Dataset 3a 120/158} \\
\hline
Algorithms & GreedyAlg &	SplitEA &	GreedyAlg &	SplitEA & GreedyAlg &	SplitEA & GreedyAlg &	SplitEA\\
\hline
K	&	78.2143 	&	\textcolor{red}{64.8381} 	&	78.2333 	&	\textcolor{red}{69.7095} 	&	135.9524 	&	\textcolor{red}{67.8381} 	&	132.2476 	&	\textcolor{red}{69.3524} 	\\
U	&	\textcolor{red}{0.3453} 	&	0.4284 	&	\textcolor{red}{0.4197} 	&	0.5104 	&	\textcolor{red}{0.4556} 	&	0.5489 	&	\textcolor{red}{0.4456} 	&	0.5272 	\\
Udelay	&	\textcolor{red}{0.1779} 	&	0.3233 	&	\textcolor{red}{0.2150} 	&	0.3218 	&	\textcolor{red}{0.0184} 	&	0.3585 	&	\textcolor{red}{0.0215} 	&	0.3341 	\\
Uunder1	&	0.1675 	&	\textcolor{red}{0.1051} 	&	0.2046 	&	\textcolor{red}{0.1886} 	&	0.4372 	&	\textcolor{red}{0.1904} 	&	0.4241 	&	\textcolor{red}{0.1932} 	\\
\hline
f	&	1.1275 	&	\textcolor{red}{1.0768} 	&	\textcolor{red}{1.2020} 	&	1.2075 	&	1.8151 	&	\textcolor{red}{1.2273} 	&	1.7681 	&	\textcolor{red}{1.2208} 	\\
\hline
\hline
Datasets & \multicolumn{2}{|c|}{Dataset 1c-Milan} &\multicolumn{2}{|c|}{Dataset 1c-Songliao} & \multicolumn{2}{c|}{Dataset 2b-Np=10 (Nt=174)} & \multicolumn{2}{c|}{Dataset 2b-Np=5 (Nt=185)} \\
\hline
Algorithms & GreedyAlg &	SplitEA &	GreedyAlg &	SplitEA & GreedyAlg &	SplitEA & GreedyAlg &	SplitEA\\
\hline
K	&	60.8381 	&	\textcolor{red}{46.5810} 	&	56.3381 	&	\textcolor{red}{46.2476} 	&	70.6476 	&	\textcolor{red}{45.9190} 	&	76.4095 	&	\textcolor{red}{56.2286} 	\\
U	&	0.6018 	&	\textcolor{red}{0.5326} 	&	0.6585 	&	\textcolor{red}{0.6049} 	&	\textcolor{red}{0.2093} 	&	0.3738 	&	\textcolor{red}{0.0707} 	&	0.4131 	\\
Udelay	&	\textcolor{red}{0.0043} 	&	0.0315 	&	\textcolor{red}{0.0044} 	&	0.0153 	&	\textcolor{red}{0.0297} 	&	0.3396 	&	\textcolor{red}{0.0438} 	&	0.3982 	\\
Uunder1	&	0.5975 	&	\textcolor{red}{0.5011} 	&	0.6541 	&	\textcolor{red}{0.5896} 	&	0.1796 	&	\textcolor{red}{0.0342} 	&	0.0270 	&	\textcolor{red}{0.0149} 	\\
\hline
f	&	1.2102 	&	\textcolor{red}{0.9985} 	&	1.2219 	&	\textcolor{red}{1.0673} 	&	0.9158 	&	\textcolor{red}{0.8329} 	&	\textcolor{red}{0.8348} 	&	0.9753 	\\
\hline
\end{tabular}
\vspace{-0.48cm}
\end{center}
\begin{tablenotes}
      \footnotesize
      \item There are 30 independent runs. The values in this table are the mean value of the metrics under 30 runs. Friedman and Nemenyi statistical tests \cite{demvsar2006statistical} with the significance level 0.05 are used to indicate the statistical significance between compared algorithms. The metric value obtained by a given algorithm on one dataset is regarded as an observation to compose that algorithm’s group for the test, following Demsar’s guidelines \cite{demvsar2006statistical}. Therefore, there are 30 observations in each group for each metric on each dataset. The significantly better values obtained by the algorithm are highlighted in red color.
    \end{tablenotes}
\end{table}

\begin{figure}[!t]
  \centering
  \subfigure[Dataset 1a]{
    \includegraphics[width=.487\textwidth]{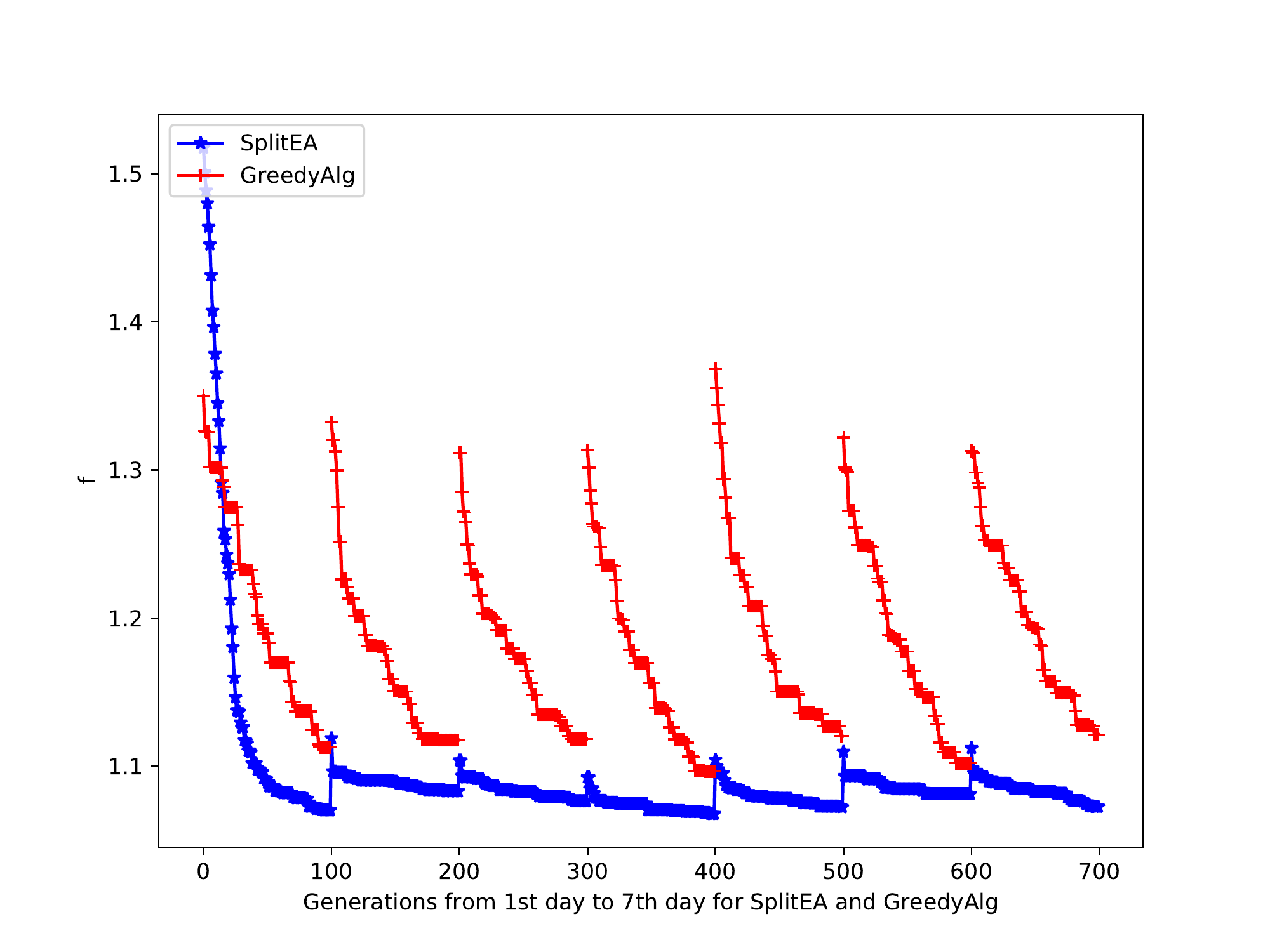}}
  \subfigure[Dataset 3a 100/158]{
    \includegraphics[width=.487\textwidth]{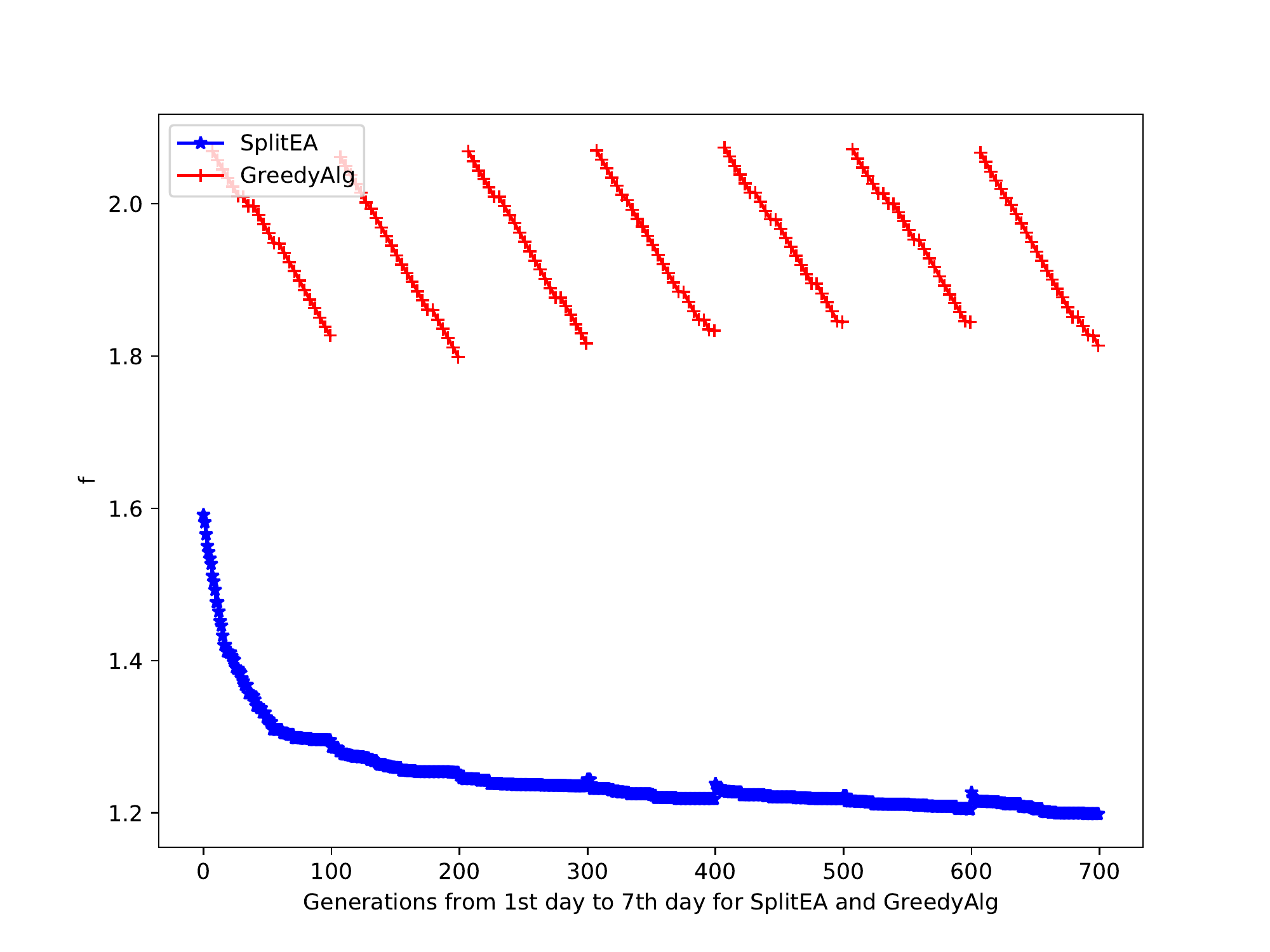}}
    \vspace{-0.4cm}
    \subfigure[Dataset 1c-Milan]{
    \includegraphics[width=.487\textwidth]{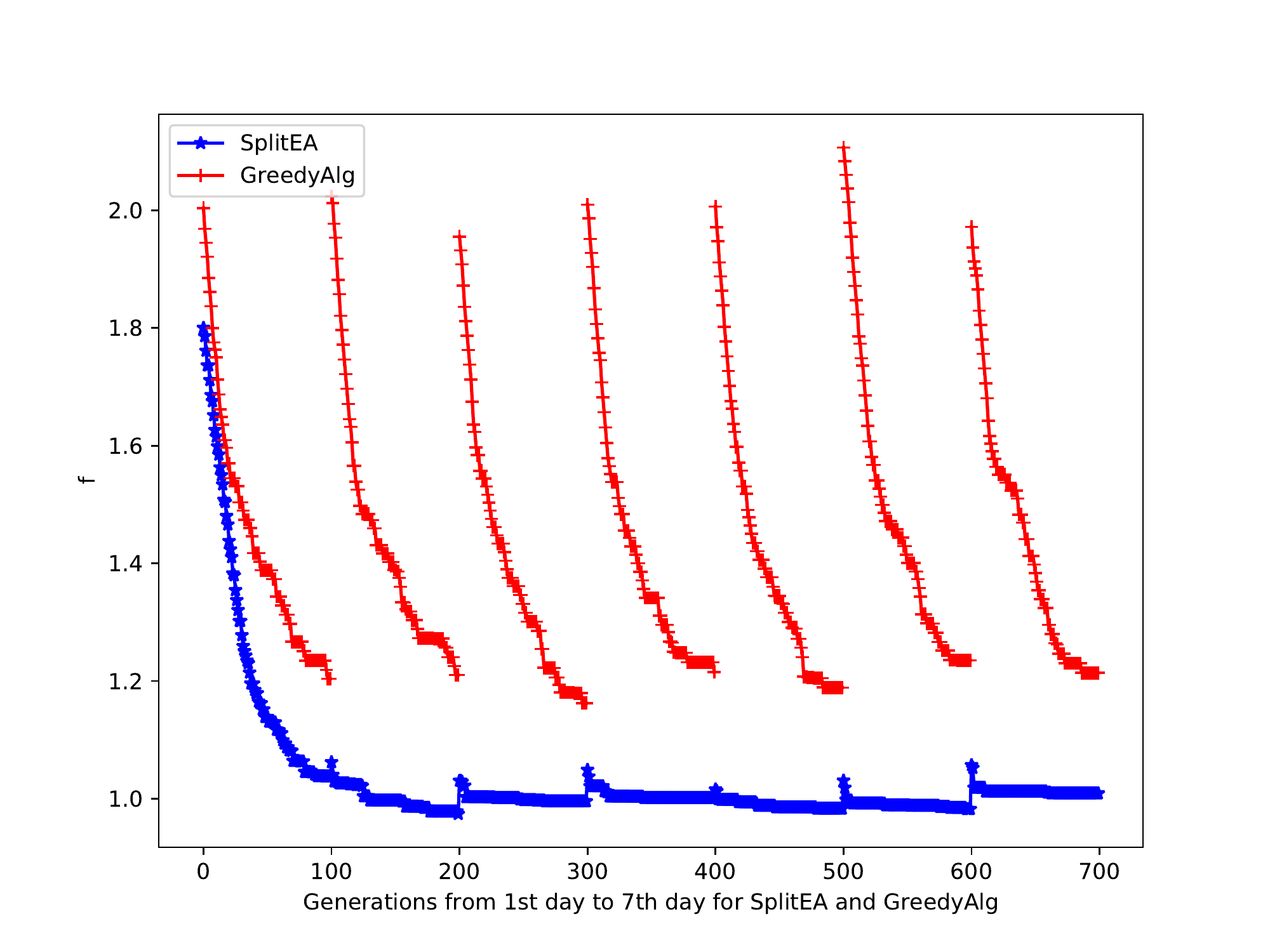}}
  \subfigure[Dataset 2b-Np=10 (Nt=174)]{
    \includegraphics[width=.487\textwidth]{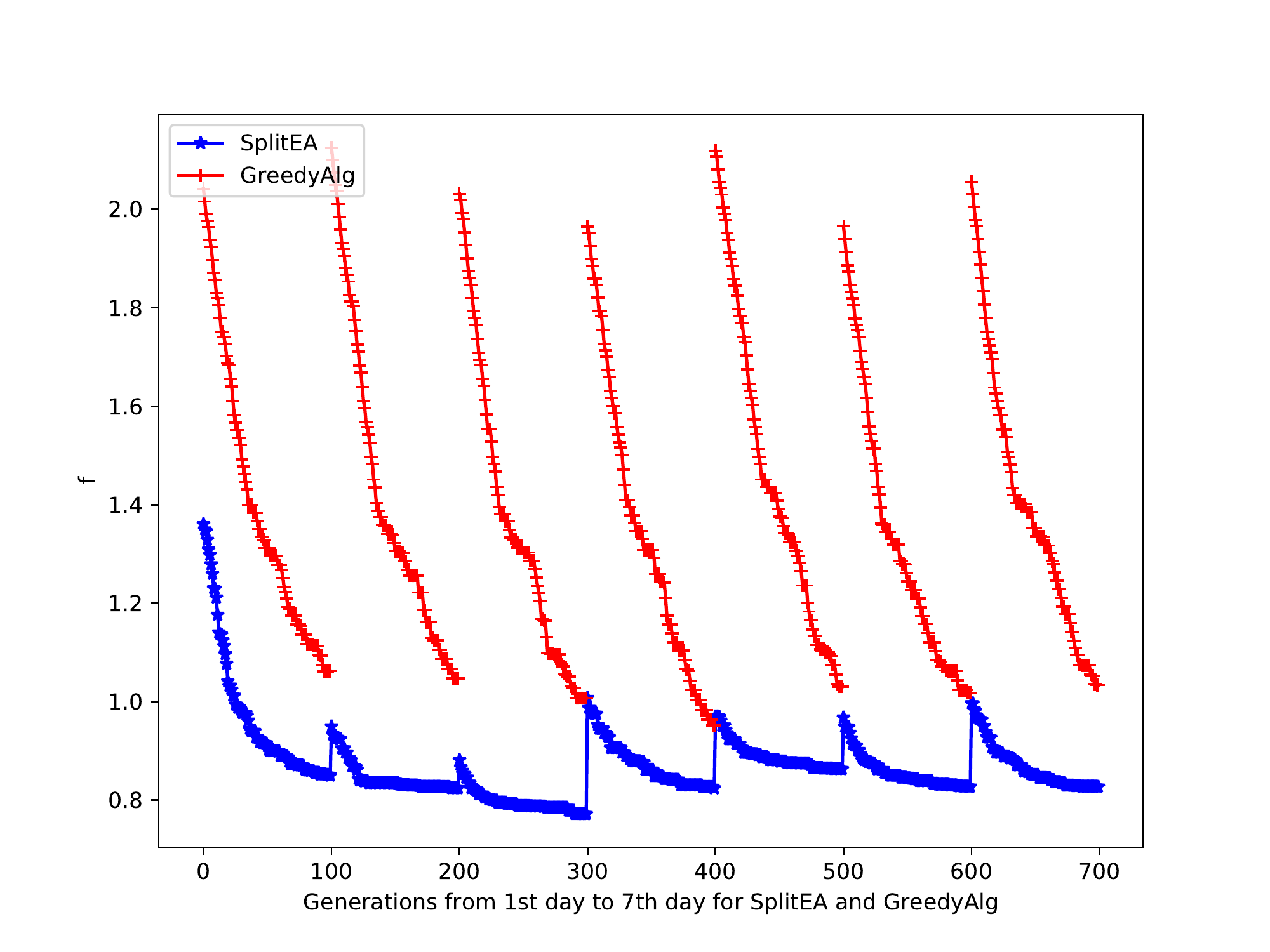}}
  \caption{Curve of the fitness value obtained by SplitEA and the greedy algorithm on four artificial datasets across the whole evolution process.}
  \label{fig:curveSplitGreedy} 
\end{figure}

\paragraph{Comparison results of SplitEA and greedy algorithm on artificial datasets}In order to test the ability of the SplitEA in searching for good solutions over the greedy algorithm on datasets with more properties beyond those of the existing datasets, like different types of location information and traffic patterns, six artificial datasets with different points distribution and different traffic pattern have been generated to do the verification. The property of those datasets are described in \ref{sec:datasets}. The comparison results of the greedy algorithm and SplitEA on the artificial predicted traffic dataset are shown in Table \ref{tab:GreedySplitArtif}. It is clear from this table that SplitEA significantly performs better than the greedy algorithm regarding the fitness values on all artificial datasets except for datasets $2a$ and $2b-Np=5 (Nt=185)$, which proves that the greedy algorithm is able to find better solutions than the greedy algorithm. As for the specific metrics, SplitEA gets significant better values of all metrics except for $Udelay$ on two datasets $1c-Milan$ and $1c-Songliao$. On other datasets, SplitEA gets significantly better $K$ and $Uunder1$ while worse $U$ and $Udelay$. 

The reasons are analyzed as follows. It is intuitive to get the reason why SplitEA gets worse $Udelay$ that SplitEA tends to cluster more points together due to the less required number of clusters for fixed number of points and this would inevitably increase the delay in the network. As for the reason why SplitEA gets worse $U$ on all datasets except for $1c-Milan$ and $1c-Songliao$, it is because the traffic datasets for all other datasets except for $1c-Milan$ and $1c-Songliao$ are artificially generated, which results in much larger mean traffic value than that of real traffic datasets. In this case, when SplitEA clusters more points together, it would get much more $Udelay$, further increasing the value of $U$. As for the reason why SplitEA gets worse fitness value on datasets $2a$ and $2b-Np=5 (Nt=185)$, mutation process tends to cluster points in each group together on the artificial datasets with the second location dataset especially when the maximal number of points in each group is small. This is the reason why SplitEA gets better fitness value on dataset $2b-Np=10 (Nt=174)$ where maximal number of points in each group is 10.




\paragraph{Fitness value curve of SplitEA and greedy algorithm on selected artificial datasets}

In order to have a better understanding of the behavior of SplitEA against the greedy algorithm in the optimization process over time, the curve of the fitness value obtained by SplitEA and greedy algorithm on four representative artificial datasets (1a, 3a 100/158, 1c-Milan and 2b-Np=10 (Nt=174)) are drawn to be shown in Figure \ref{fig:curveSplitGreedy}. In other words, we need to check if the better performance of the proposed SplitEA than that of the greedy algorithm is consistent over time. It is clear from Figure \ref{fig:curveSplitGreedy} that SplitEA gets better fitness values at all days of dataset $3a 100/158$ and dataset $2b-Np=10 (Nt=174)$. In addition, SplitEA only gets worse fitness value at the beginning of the first day of dataset $1a$ and dataset $1c-Milan$. Therefore, it has been checked that there are not some periods of time for which SplitEA is actually much worse than the greedy algorithm, despite being better overall. To be more specific, SplitEA is able to get better fitness values than the greedy algorithm in the optimization process over time.


\paragraph{The effect of different parameter settings on the performance of SplitEA and greedy algorithm}

In order to check the influence of different parameter settings on the performance of SplitEA and greedy algorithm, two methods are tested on the Milan Dataset which sets different values for two problem-related parameters $w$ and $\tau$. The comparison results of SplitEA and greedy algorithm on dataset Milan under different parameter settings are presented in Table \ref{tab:ParaGreedySplit}. It is clear from this table that under different settings of $w$, SplitEA significantly performs better than the greedy algorithm on all metrics and the fitness value. In addition, SplitEA gets significantly better results the greedy algorithm on all metrics under all settings of $\tau$ except for the setting of $\tau=1500$ on the metric $Udelay$. The reason might be that if $\tau$ is too large, SplitEA tends to cluster many points together, which inevitably causes more delay than the greedy algorithm causes.

\begin{table}[ht]
\begin{center}
\caption{The effect of parameters $w$ and $\tau$ in the problem on the performance of SplitEA and greedy algorithm.}
\label{tab:ParaGreedySplit}
\begin{tabular}{|c|cc|cc|cc|}
\hline
w & \multicolumn{2}{|c|}{0.001} &\multicolumn{2}{|c|}{0.01} & \multicolumn{2}{c|}{1} \\
\hline
Algorithms & GreedyAlg &	SplitEA &	GreedyAlg &	SplitEA & GreedyAlg &	SplitEA\\
\hline
K&	61.1725 &	\textcolor{red}{53.2882} &	61.1863 &	\textcolor{red}{52.0980} &	59.9784 &	\textcolor{red}{51.7137} \\
U	&0.8326 &	\textcolor{red}{0.7623} &	0.8323 &	\textcolor{red}{0.7659} 	&0.8295 	&\textcolor{red}{0.7684} \\
Udelay&	0.0236 &	\textcolor{red}{0.0026} &	0.0238 &	\textcolor{red}{0.0077} &	0.0244 	&\textcolor{red}{0.0100} \\
Uunder1	&0.8090 &	\textcolor{red}{0.7596} &	0.8086 &	\textcolor{red}{0.7582} &	0.8051 &	\textcolor{red}{0.7583}\\ 
f&	0.8937 &	\textcolor{red}{0.8156} &	1.4442 &	\textcolor{red}{1.2869}& 	60.8079& 	\textcolor{red}{52.4821}\\
\hline
\hline
$\tau$ & \multicolumn{2}{|c|}{800} &\multicolumn{2}{|c|}{1000} & \multicolumn{2}{c|}{1500} \\
\hline
Algorithms & GreedyAlg &	SplitEA &	GreedyAlg &	SplitEA & GreedyAlg &	SplitEA\\
\hline
K	&68.9020 	&\textcolor{red}{62.6255} 	&59.9784 	&\textcolor{red}{51.7137} &	92.9392 	&\textcolor{red}{35.0765} \\
U	&0.8397 &	\textcolor{red}{0.8048}& 	0.8295 &	\textcolor{red}{0.7684} &	0.8823 	&\textcolor{red}{0.6681} \\
Udelay	&0.0151 	&\textcolor{red}{0.0064} 	&0.0244 &	\textcolor{red}{0.0100} 	&\textcolor{red}{0.0122} 	&0.0185 \\
Uunder1&	0.8246 	&\textcolor{red}{0.7983}& 	0.8051 	&\textcolor{red}{0.7583} 	&0.8702 &	\textcolor{red}{0.6496} \\
f	&69.7416 &	\textcolor{red}{63.4303}& 	60.8079& 	\textcolor{red}{52.4821} &	93.8215 &	\textcolor{red}{35.7446}\\
\hline
\end{tabular}
\vspace{-0.48cm}
\end{center}
\begin{tablenotes}
      \footnotesize
      \item There are 30 independent runs. The values in this table are the mean value of the metrics under 30 runs. Friedman and Nemenyi statistical tests \cite{demvsar2006statistical} with the significance level 0.05 are used to indicate the statistical significance between compared algorithms. The metric value obtained by a given algorithm on one dataset is regarded as an observation to compose that algorithm’s group for the test, following Demsar’s guidelines \cite{demvsar2006statistical}. Therefore, there are 30 observations in each group for each metric on each parameter setting. The significantly better values obtained by the algorithm are highlighted in red color.
    \end{tablenotes}
\end{table}

\subsubsection{Analyses of Random Cluster Splitting in SplitEA}
\label{sec:ComEAs}

This section tries to answer the second sub-research question of the second question: What is the influence of different algorithm’s design choices on its performance? Considering that process of random cluster splitting is an importance mechanism in SplitEA and there are two intuitive alternatives, this section analyzes the effect of the process of random cluster splitting on SplitEA. Therefore, we replace the process of random cluster splitting with two different processes to create two variants of SplitEA, named as RandEA and CopyEA, and compare three of them on several real-world and artificial datasets. The details of the replacement in RandEA and CopyEA can be found in Section \ref{sec:comAlgo}. Firstly, SplitEA and two variants are tested on the real-world datasets to verify that whether the proposed SplitEA gets best optimization results than two variants on real-world datasets. Then, three algorithms are tested on predicted traffic datasets to see whether the prediction error affects the performance of SplitEA. 
Lastly, three compared algorithms are tested on artificial datasets with more properties beyond the existing datasets. Note for those experiments without using the prediction model, the Step 6 in Algorithm \ref{Alg:framework} is deleted and in Step 7, the fitness value of solutions is calculated on the real traffic data.

\begin{table}[ht]
\scriptsize
\begin{center}
\caption{Comparison results of three EAs on real-world datasets.}
\label{tab:EAsReal}
\begin{tabular}{|c|ccc|ccc|ccc|}
\hline
Datasets & \multicolumn{3}{|c|}{Milan} &\multicolumn{3}{|c|}{Songliao} & \multicolumn{3}{c|}{C2TM-sub1} \\
\hline
Algorithms & RandEA & CopyEA &	SplitEA &	RandEA & CopyEA &	SplitEA & RandEA & CopyEA &	SplitEA\\
\hline
K	&55.5481 &	\textbf{53.3444}& 	\textcolor{red}{52.3019}& 	\textcolor{red}{53.4524} &	\textbf{53.8810} &	54.3619 &	\textbf{54.3792} &	\textcolor{red}{52.0458}& 	\textcolor{red}{51.7708} \\
U	&0.7766 &	\textbf{0.7671} &	\textcolor{red}{0.7644} &	0.8187 &	\textbf{0.8109} &	\textcolor{red}{0.8042} &	\textbf{0.9958} &	\textcolor{red}{0.9956} &	\textcolor{red}{0.9956}\\ 
Udelay&	0.0067 &	0.0069 &	0.0076 &	0.0351 &	\textbf{0.0274} &	\textcolor{red}{0.0234} &	1.70E-07	&2.84E-07	&2.97E-07\\
Uunder1&	0.7699 &	\textbf{0.7602} &	\textcolor{red}{0.7568} &	0.7836 &	\textbf{0.7778} &	\textcolor{red}{0.7760} &	\textbf{0.9958} &	\textcolor{red}{0.9956} &	\textcolor{red}{0.9956} \\
\hline
f	&1.3321 &	\textbf{1.3005} &	\textcolor{red}{1.2874} &	1.3532 &	\textbf{1.3497} &	\textcolor{red}{1.3478} &	\textbf{1.5396} &	\textcolor{red}{1.5160} &	\textcolor{red}{1.5133}\\
\hline
\hline
Datasets & \multicolumn{3}{|c|}{Archive-Milan} &\multicolumn{3}{|c|}{Archive-Songliao} & \multicolumn{3}{c|}{Archive-Random} \\
\hline
Algorithms & RandEA & CopyEA &	SplitEA &	RandEA & CopyEA &	SplitEA & RandEA & CopyEA &	SplitEA\\
\hline
K	&13.5239 &	\textbf{13.1157} &	\textcolor{red}{13.0352} &	13.5308& 	\textcolor{red}{12.9308}& 	\textbf{13.2629} &	\textbf{12.1497} &	\textcolor{red}{12.0057} &	\textcolor{red}{12.0069} \\
U&	0.7587 &	\textbf{0.7548} &	\textcolor{red}{0.7536} &	\textbf{0.7572} &	\textcolor{red}{0.7522} &	\textcolor{red}{0.7538} &	0.7289 &	\textbf{0.7268} &	\textcolor{red}{0.7260} \\
Udelay	&\textcolor{red}{0.0020} &	\textbf{0.0038} &	\textbf{0.0039} &	\textcolor{red}{0.0011} &	0.0043 &	\textbf{0.0022} &	\textcolor{red}{0.0008} &	\textbf{0.0014} &	\textcolor{red}{0.0009} \\
Uunder1&	0.7567& 	\textbf{0.7510} &	\textcolor{red}{0.7497} &	\textbf{0.7560}&	\textcolor{red}{0.7479} &	\textcolor{red}{0.7517} &	0.7280 &	\textbf{0.7254} &	\textcolor{red}{0.7250} \\
\hline
f	&0.8939 &	\textbf{0.8859} &	\textcolor{red}{0.8840} &	0.8925 &	\textcolor{red}{0.8815} &	\textbf{0.8865} &	0.8504 &	\textbf{0.8469} &	\textcolor{red}{0.8460}\\
\hline
\end{tabular}
\vspace{-0.48cm}
\end{center}
\begin{tablenotes}
      \footnotesize
      \item There are 30 independent runs. The values in this table are the mean value of the metrics under 30 runs. The best and the second best values obtained by the algorithm are highlighted in red and bold face, respectively. Friedman and Nemenyi statistical tests \cite{demvsar2006statistical} with the significance level 0.05 are used to indicate the statistical significance between compared algorithms. The metric value obtained by a given algorithm on one dataset is regarded as an observation to compose that algorithm’s group for the test, following Demsar’s guidelines \cite{demvsar2006statistical}. Therefore, there are 30 observations in each group for each metric on each dataset. If results obtained by two or three algorithms are highlighted with the same mark, it meas there is no significant difference between them.
    \end{tablenotes}
\end{table}

\paragraph{Comparison results of SplitEA and two variants on real-world datasets}
The comparison results of three algorithms on the real-world dataset are shown in Table \ref{tab:EAsReal}. It is clear that SplitEA gets significant overall best results on all datasets except for the dataset Archive-Songliao. The reason why CopyEA gets significant better results than SplitEA on the dataset Archive-Songliao is due to the location dataset of Archive-Songliao, as three datasets (Archive-Milan, Archive-Songliao and Archive-Random) have the same traffic dataset. It is possible on this dataset that when the solutions from the problem of the previous day are still good for the problem of the next day. Therefore, when SplitEA splits one cluster into two clusters, it makes the solutions worse. On all datasets except for Archive-Songliao, SplitEA gets significantly best values on three metrics. Specifically, SplitEA is the best regarding $K$, $U$ and $Uunder1$ on datasets Milan and $Archive-Milan$. It is easy to understand that when SplitEA gets the smallest number of clusters while resulting more delay. As for the reason why CopyEA and SplitEA perform equally on C2TM, it is due to the traffic dataset, which is so few that the SplitEA could not find better solutions than CopyEA. With regards to the results on dataset Songliao, SplitEA gets the worse $K$ while RandEA gets the best. It might be that when further increasing the resource utilization rate and decreasing the delay, the number of clusters inevitably increases.

\begin{table}[ht]
\tiny
\begin{center}
\caption{Comparison results of three EAs on real-world datasets with prediction.}
\label{tab:EAsPred}
\begin{tabular}{|c|ccc|ccc|ccc|ccc|}
\hline
Datasets & \multicolumn{3}{|c|}{Milan} &\multicolumn{3}{|c|}{Archive-Milan} & \multicolumn{3}{c|}{Archive-Songliao} & \multicolumn{3}{c|}{Archive-Random} \\
\hline
Algorithms & RandEA & CopyEA &	SplitEA &	RandEA & CopyEA &	SplitEA & RandEA & CopyEA &	SplitEA& RandEA & CopyEA &	SplitEA\\
\hline
K	&\textbf{13.3635} &	13.6233 &	\textcolor{red}{13.0340} 	&13.4692 	&\textbf{13.0962} &	\textcolor{red}{13.0181} &13.2068& 	\textbf{12.8242} &	\textcolor{red}{12.8106} &	\textbf{12.0794} &	\textcolor{red}{12.0019}& 	\textcolor{red}{12.0019} \\
U	&\textbf{0.7591} &	0.7611 &	\textcolor{red}{0.7547} &	0.7594 &	\textbf{0.7546} 	&\textcolor{red}{0.7540}&0.7608 &	\textcolor{red}{0.7535} &	\textbf{0.7553} &	0.7284 &	\textbf{0.7268} &	\textcolor{red}{0.7267} \\
Udelay	&	\textbf{0.0036} &	\textcolor{red}{0.0021} &	0.0044 &	\textcolor{red}{0.0029} &	\textbf{0.0039} 	&0.0043 &\textcolor{red}{0.0060} &	\textbf{0.0061} &	0.0072 &	\textbf{0.0014} &	\textbf{0.0014} &	\textcolor{red}{0.0013} \\
Uunder1	&	\textbf{0.7555} &	0.7590 &	\textcolor{red}{0.7502} &	0.7566 &	\textbf{0.7507} &	\textcolor{red}{0.7497}&0.7548 &	\textcolor{red}{0.7473} &	\textbf{0.7480} &	0.7270 &	\textbf{0.7254} &	\textcolor{red}{0.7253}  \\
\hline
f	&\textbf{0.8928} &	0.8973 &	\textcolor{red}{0.8850} &	0.8941 	&\textbf{0.8855} 	&\textcolor{red}{0.8842}&	0.8929 &	\textcolor{red}{0.8817} &	\textbf{0.8834} &	0.8492 &	\textbf{0.8468} &	\textcolor{red}{0.8467} \\
\hline
\end{tabular}
\vspace{-0.48cm}
\end{center}
\begin{tablenotes}
      \footnotesize
      \item There are 30 independent runs. The values in this table are the mean value of the metrics under 30 runs. The best and the second best values obtained by the algorithm are highlighted in red and bold face, respectively. Friedman and Nemenyi statistical tests \cite{demvsar2006statistical} with the significance level 0.05 are used to indicate the statistical significance between compared algorithms. The metric value obtained by a given algorithm on one dataset is regarded as an observation to compose that algorithm’s group for the test, following Demsar’s guidelines \cite{demvsar2006statistical}. Therefore, there are 30 observations in each group for each metric on each dataset. If results obtained by two or three algorithms are highlighted with the same mark, it meas there is no significant difference between them.
    \end{tablenotes}
\end{table}

\begin{table}[ht]
\tiny
\begin{center}
\caption{Comparison results of three EAs on artificial datasets.}
\label{tab:EAsArtifi}
\begin{tabular}{|c|ccc|ccc|ccc|ccc|}
\hline
Datasets & \multicolumn{3}{|c|}{Dataset 1a		
} &\multicolumn{3}{|c|}{Dataset 2a		
} & \multicolumn{3}{c|}{Dataset 3a 100/158		
} & \multicolumn{3}{c|}{Dataset 3a 120/158		
} \\
\hline
Algorithms & RandEA & CopyEA &	SplitEA &	RandEA & CopyEA &	SplitEA & RandEA & CopyEA &	SplitEA& RandEA & CopyEA &	SplitEA\\
\hline
K&	64.7429& 	64.5714 &	64.8381& 	70.5190& 	70.1429 &	69.7095& 	\textcolor{red}{61.4095} &	\textbf{63.4667}& 	67.8381 	&\textcolor{red}{63.6952} &	\textbf{66.3429} &	69.3524 \\
U	&\textbf{0.4343} &	0.4531 &	\textcolor{red}{0.4284} 	&0.5036 	&0.5127 &	0.5104 	&0.6849 	&\textbf{0.6391} 	&\textcolor{red}{0.5489} 	&0.6307 	&\textbf{0.5809} 	&\textcolor{red}{0.5272} \\
Udelay&	\textcolor{red}{0.3272} &	\textbf{0.3381} &	\textcolor{red}{0.3233} &	\textcolor{red}{0.3120} 	&\textbf{0.3195} &	0.3218 &	0.4858 &	\textbf{0.4425} &	\textcolor{red}{0.3585} 	&0.4356 	&\textbf{0.3860} &	\textcolor{red}{0.3341} \\
Uunder1	&\textcolor{red}{0.1072} 	&\textbf{0.1149} &	\textcolor{red}{0.1051} &	\textbf{0.1916} 	&\textbf{0.1932 }	&\textcolor{red}{0.1886} &	0.1991 	&\textbf{0.1966} 	&\textcolor{red}{0.1904} 	&\textbf{0.1951} 	&\textbf{0.1949} 	&\textcolor{red}{0.1932} \\
\hline
f	&\textbf{1.0818 }	&1.0988 &	\textcolor{red}{1.0768} &	\textcolor{red}{1.2088} &	\textbf{1.2141} &	\textcolor{red}{1.2075} &	1.2990 &	\textbf{1.2738} &	\textcolor{red}{1.2273} 	&1.2677 	&\textbf{1.2443} &	\textcolor{red}{1.2208}\\
\hline
\hline
Datasets & \multicolumn{3}{|c|}{Dataset 1c-Milan		
} &\multicolumn{3}{|c|}{Dataset 1c-Songliao		
} & \multicolumn{3}{c|}{Dataset 2b-Np=10 (d=174)		
} & \multicolumn{3}{c|}{Dataset 2b-Np=5 (d=185)		
} \\
\hline
Algorithms & RandEA & CopyEA &	SplitEA &	RandEA & CopyEA &	SplitEA & RandEA & CopyEA &	SplitEA& RandEA & CopyEA &	SplitEA\\
\hline

K	&48.5571 &	\textbf{47.1667} 	&\textcolor{red}{46.5810} 	&\textbf{48.2952} &	\textbf{48.4571}& 	\textcolor{red}{46.2476} &	\textbf{49.0857} &	\textbf{49.6571} &	\textcolor{red}{45.9190} 	&\textbf{61.5905} 	&\textbf{61.4000} &	\textcolor{red}{56.2286} \\
U	&\textbf{0.5466} 	&0.5566 	&\textcolor{red}{0.5326} 	&\textbf{0.6193} 	&0.6388 	&\textcolor{red}{0.6049} 	&\textcolor{red}{0.3149} 	&0.4777 	&\textbf{0.3738} 	&\textcolor{red}{0.3072} 	&\textbf{0.4051} 	&\textbf{0.4131} \\
Udelay	&\textcolor{red}{0.0277} 	&0.0403 	&\textbf{0.0315} 	&\textcolor{red}{0.0135} 	&\textbf{0.0226} 	&\textcolor{red}{0.0153} 	&\textcolor{red}{0.2678} 	&\textbf{0.3421} 	&\textbf{0.3396} 	&\textcolor{red}{0.2839} 	&\textbf{0.3350} 	&0.3982 \\
Uunder1	&\textbf{0.5189} 	&\textbf{0.5163} 	&\textcolor{red}{0.5011} 	&\textbf{0.6058} &	0.6161 	&\textcolor{red}{0.5896} 	&\textbf{0.0471} 	&0.1356 	&\textcolor{red}{0.0342} 	&\textbf{0.0233} 	&0.0701 	&\textcolor{red}{0.0149} \\
\hline
f	&\textbf{1.0322} &	\textbf{1.0283} 	&\textcolor{red}{0.9985} &	\textbf{1.1022} &	1.1233 &	\textcolor{red}{1.0673} &	\textcolor{red}{0.8057} &	0.9743 	&\textbf{0.8329} 	&\textcolor{red}{0.9231} 	&1.0191 	&\textbf{0.9753}\\
\hline
\end{tabular}
\vspace{-0.48cm}
\end{center}
\begin{tablenotes}
      \footnotesize
      \item There are 30 independent runs. The values in this table are the mean value of the metrics under 30 runs. The best and the second best values obtained by the algorithm are highlighted in red and bold face, respectively. Friedman and Nemenyi statistical tests \cite{demvsar2006statistical} with the significance level 0.05 are used to indicate the statistical significance between compared algorithms. The metric value obtained by a given algorithm on one dataset is regarded as an observation to compose that algorithm’s group for the test, following Demsar’s guidelines \cite{demvsar2006statistical}. Therefore, there are 30 observations in each group for each metric on each dataset. If results obtained by two or three algorithms are highlighted with the same mark, it meas there is no significant difference between them.
    \end{tablenotes}
\end{table}

\begin{figure}[!t]
\centering
\subfigure{
\includegraphics[width=.5\textwidth]{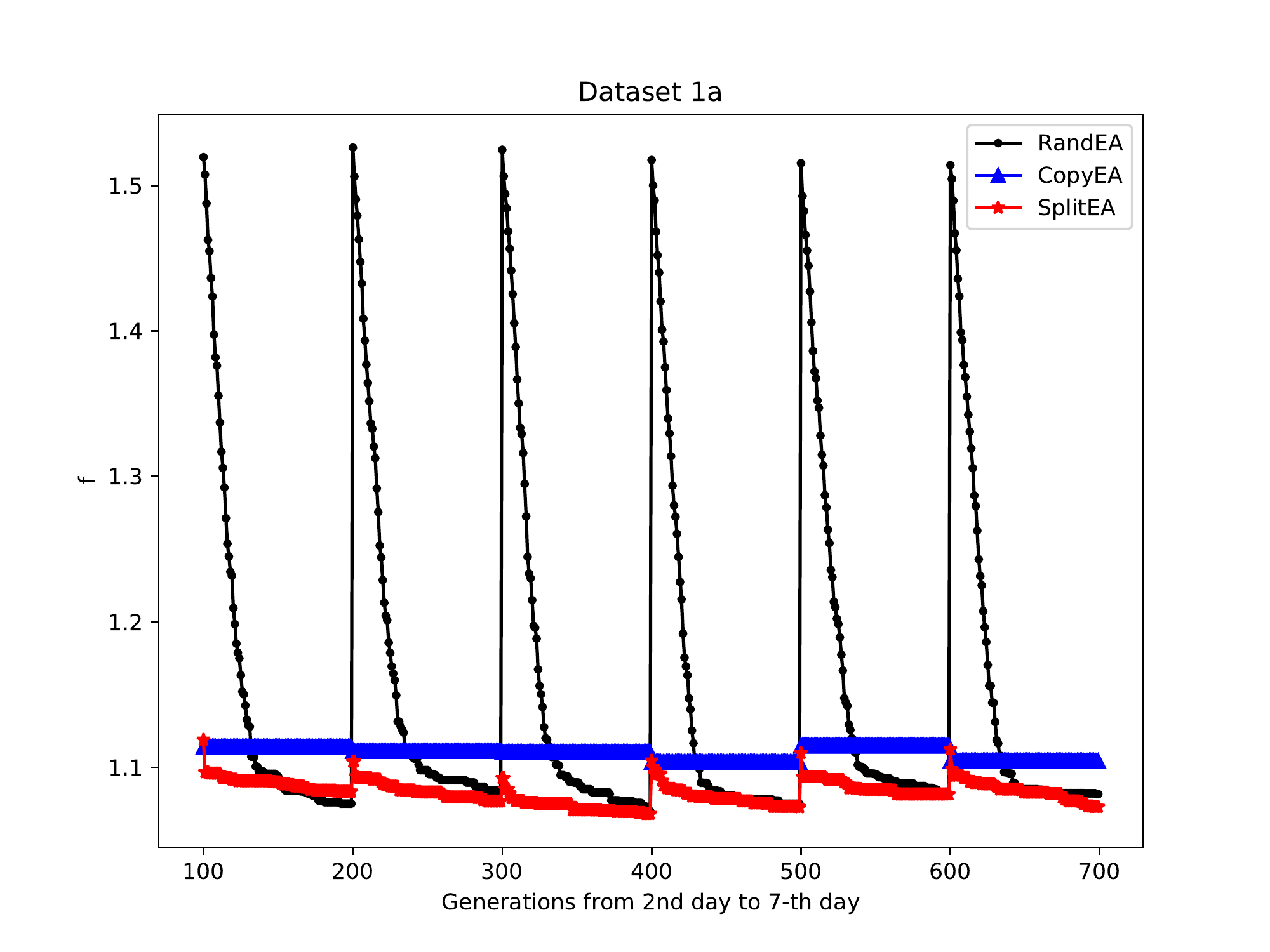}
\includegraphics[width=.5\textwidth]{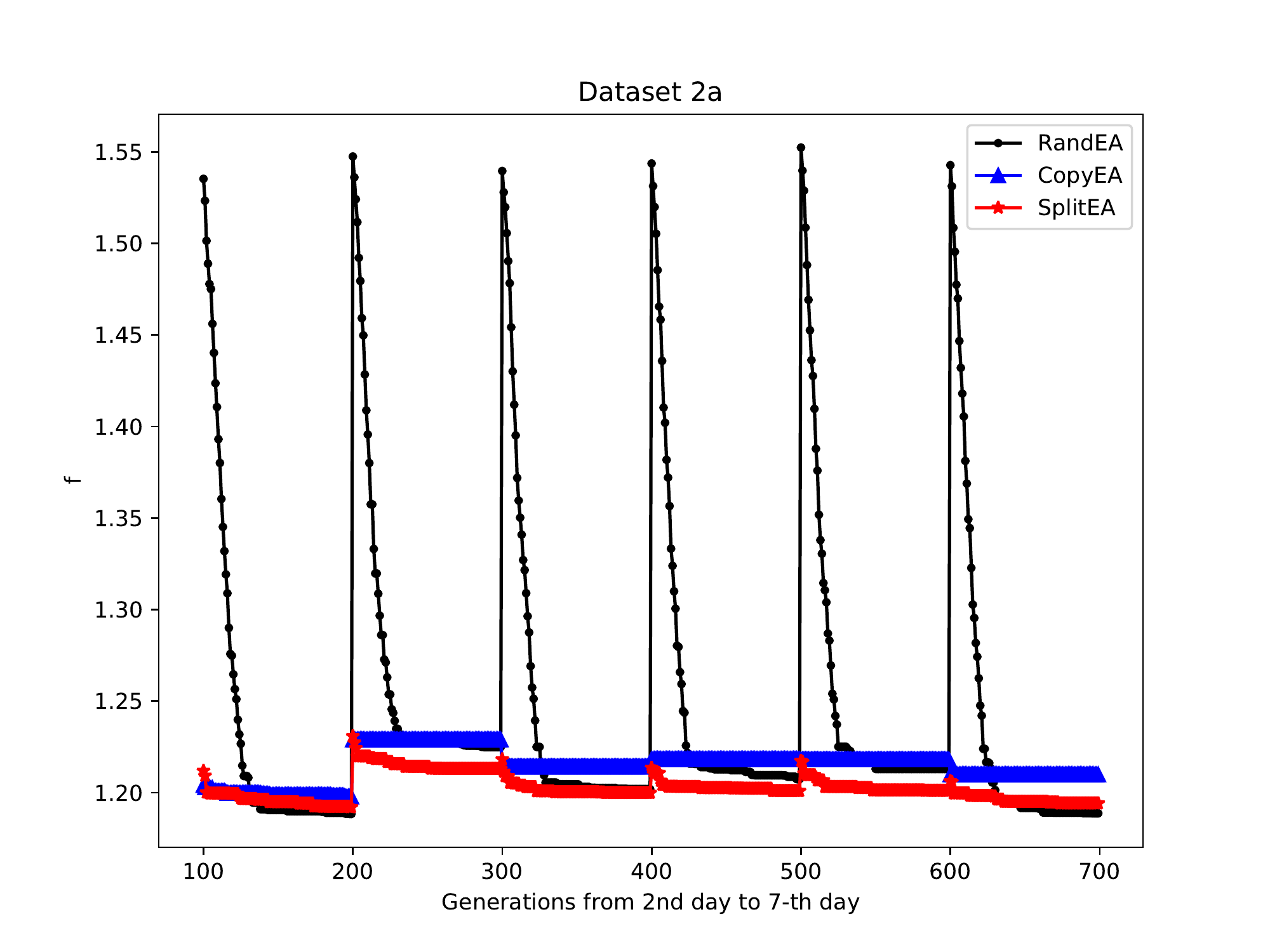}
}
\vspace{-0.3cm}
\subfigure{
\includegraphics[width=.5\textwidth]{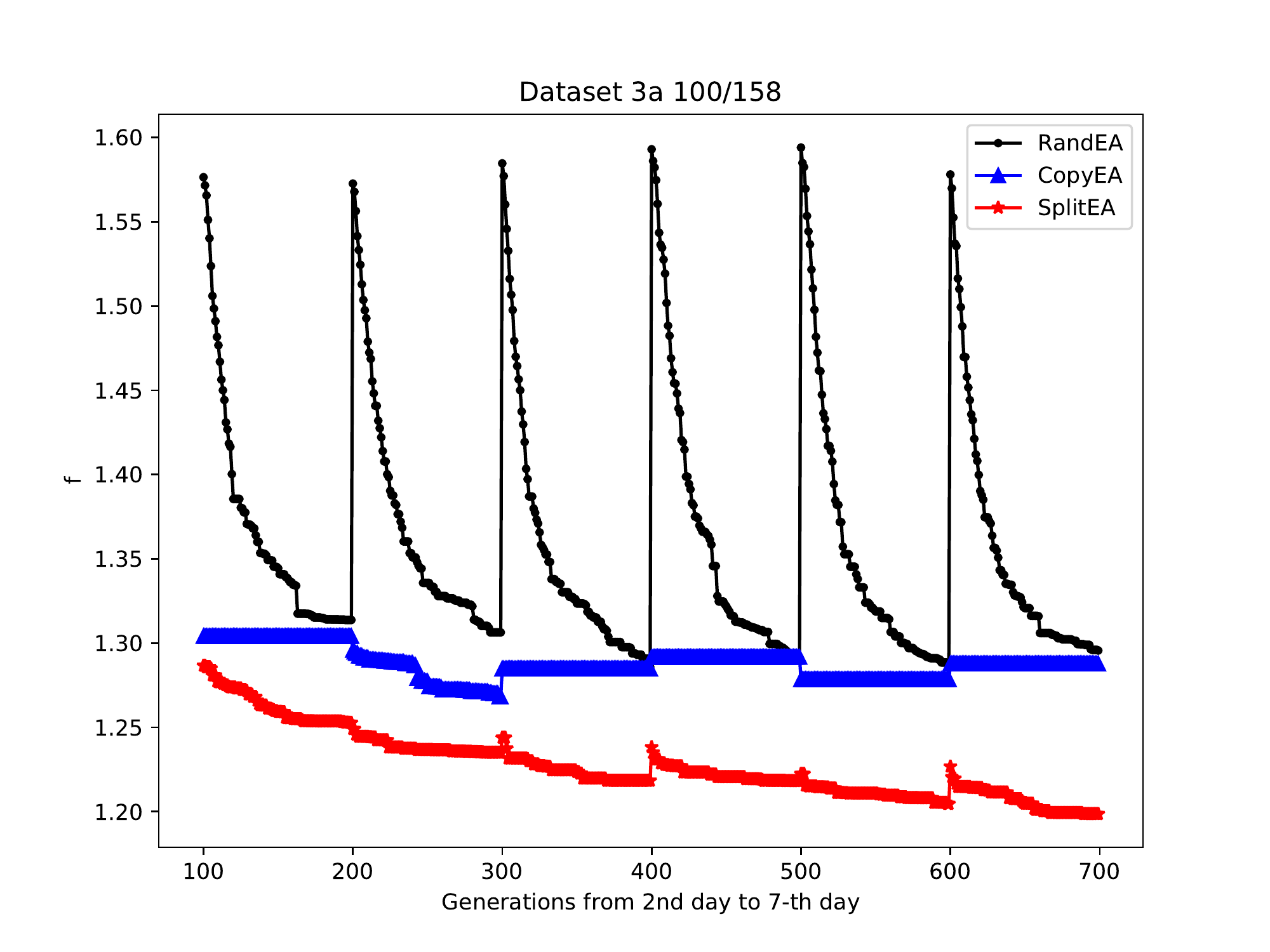}
\includegraphics[width=.5\textwidth]{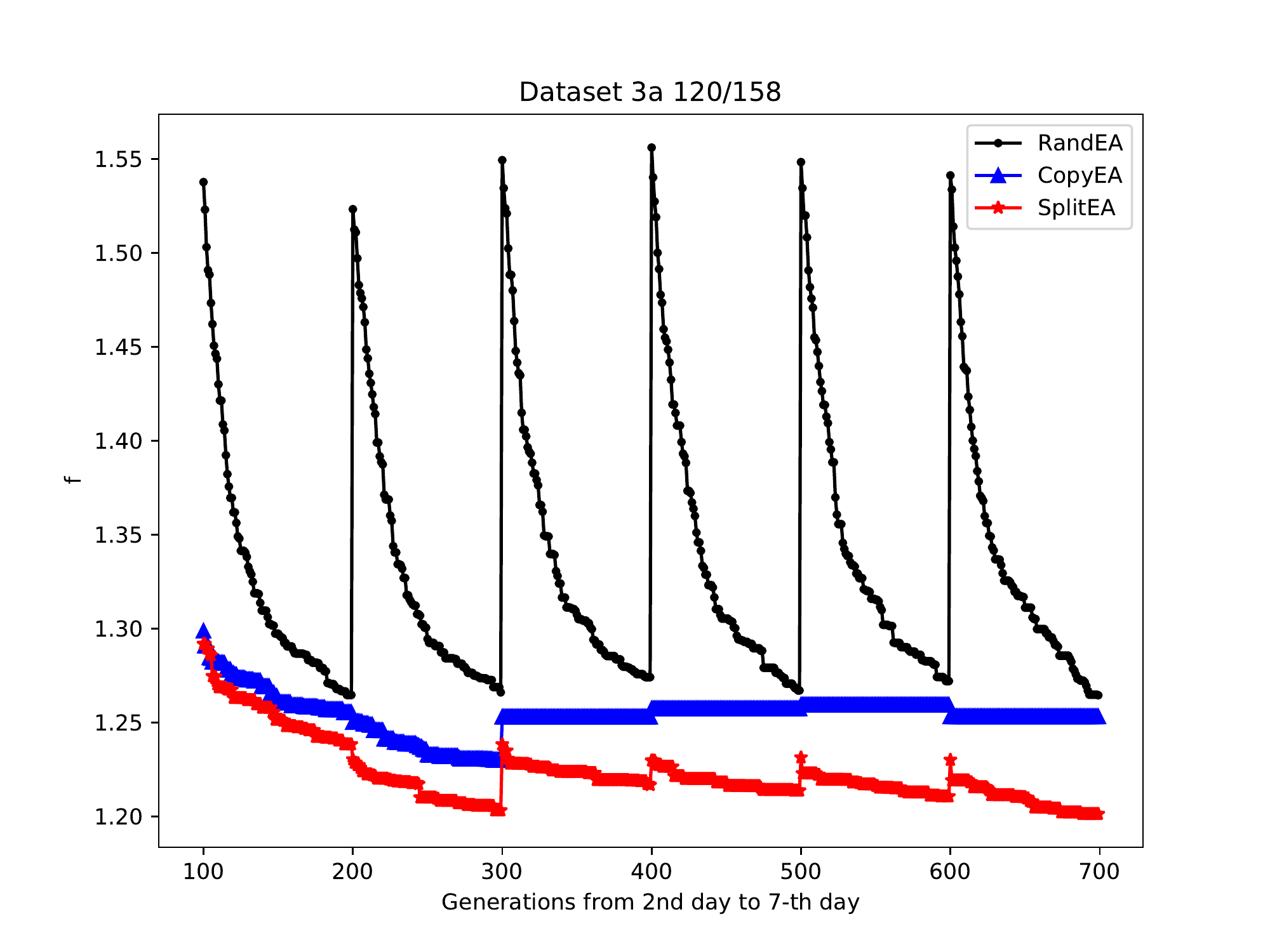}}
\caption{Curve of the fitness value obtained by SplitEA and two variants on four artificial datasets across the whole evolution process.}
\label{fig.EAsCurve1}
\end{figure}

\paragraph{Comparison results of SplitEA and two variants on datasets with prediction}
The comparison results of three algorithms on the real-world datasets with prediction are shown in Table \ref{tab:EAsPred}. It is clear that SplitEA gets the best fitness values on all four dataset except the Archive-Songliao, which is similar to those without prediction. The reason might be similar, due to the location dataset of Archive-Songliao. In addition,
SplitEA significantly performs best on datasets Milan and Archive-Milan regarding all four metrics except for the $Udelay$. The only difference between their performance on Milan with prediction and without prediction is that RandEA performs better than CopyEA on Milan with prediction. The performance of three algorithms on Archive-Random is similar to that without prediction. Therefore, it can be conclude that the prediction error does not have much influence on the performance of SplitEA.

\begin{figure}[!t]
\centering
\subfigure{
\includegraphics[width=.5\textwidth]{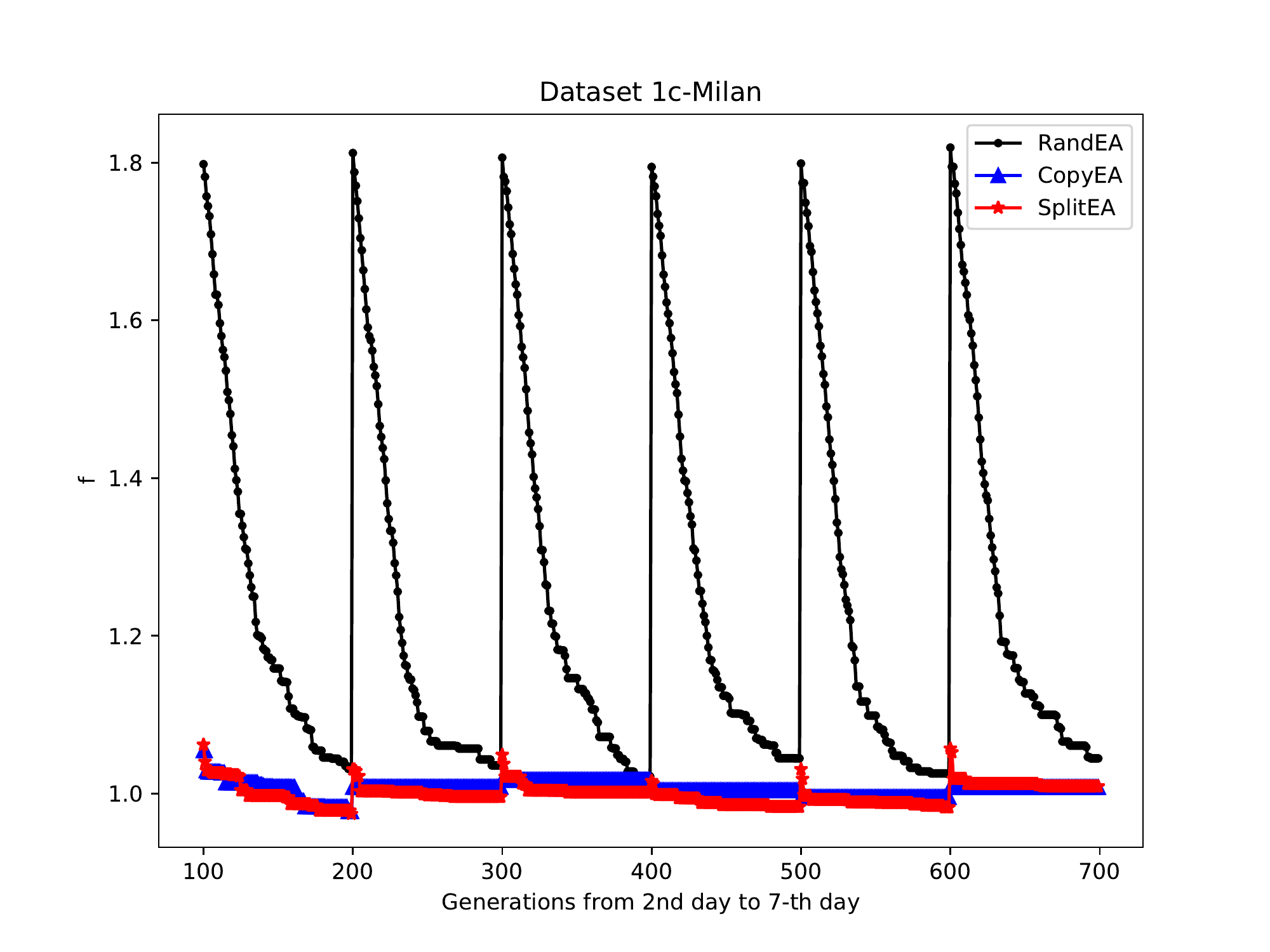}
\includegraphics[width=.5\textwidth]{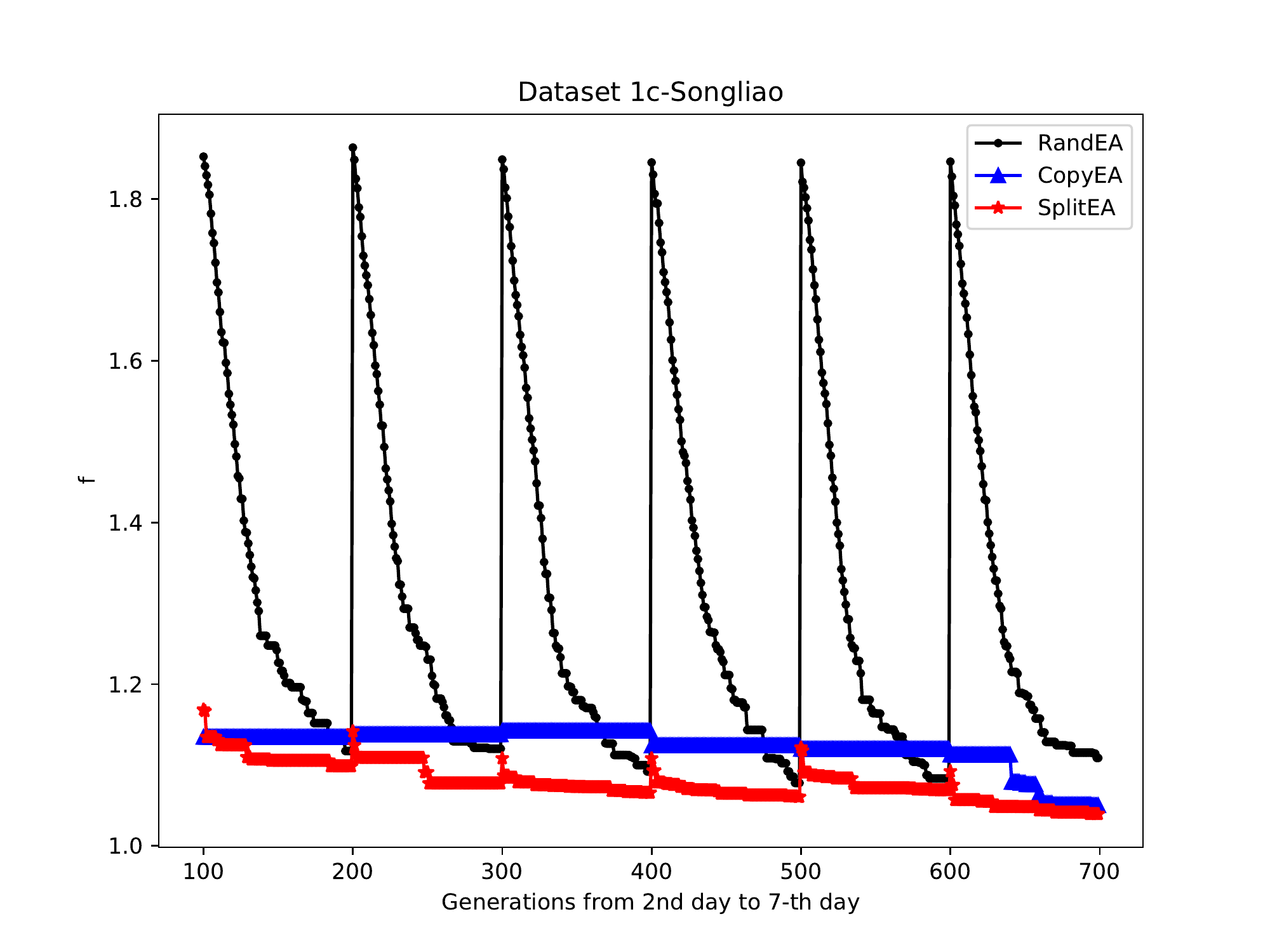}
}\vspace{-0.3cm}
\subfigure{
\includegraphics[width=.5\textwidth]{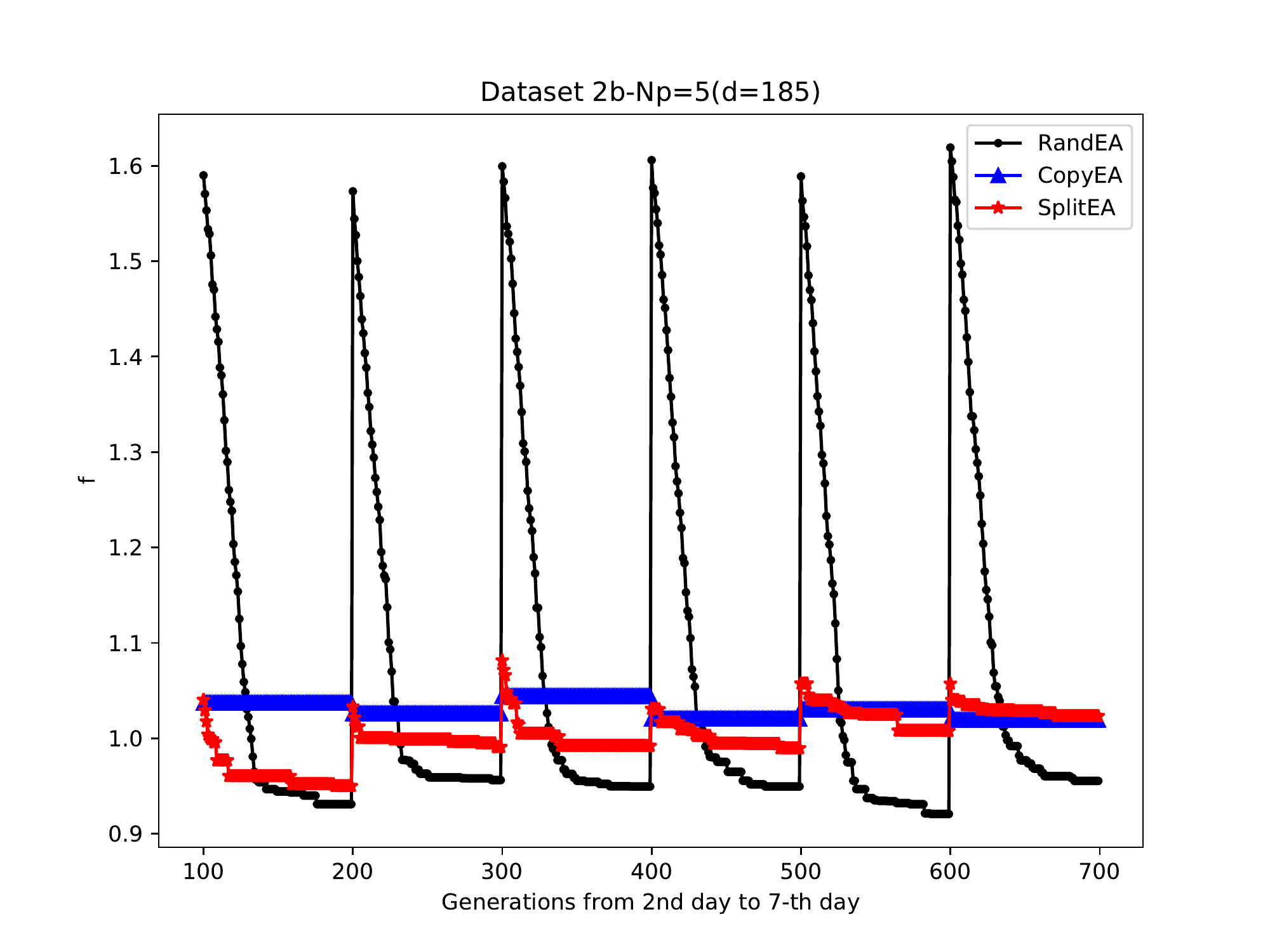}
\includegraphics[width=.5\textwidth]{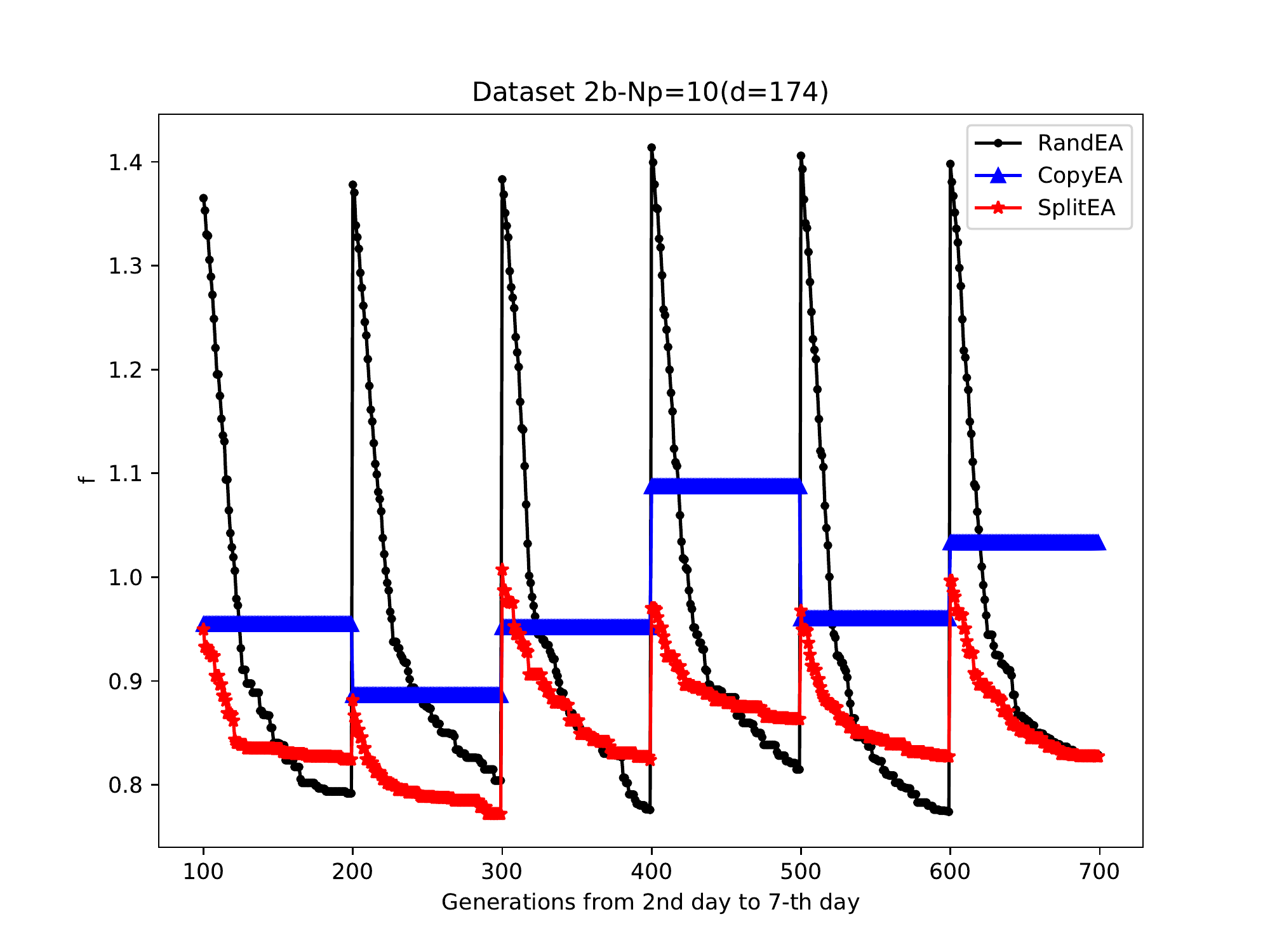}}
\caption{Curve of the fitness value obtained by SplitEA and two variants on four artificial datasets across the whole evolution process.}
\label{fig.EAsCurve2}
\end{figure}
\paragraph{Comparison results of SplitEA and two variants on artificial datasets}
In order to test the performance of the SplitEA in searching for good solutions over RandEA and CopyEA on different scenarios with different points distribution and different traffic pattern, several artificial dataset have been generated to do the verification. The property of those datasets are described in \ref{sec:datasets}. The comparison results of the greedy algorithm and SplitEA on the artificial predicted traffic dataset are shown in Table \ref{tab:EAsArtifi}. It is clear from this table that SplitEA significantly performs best than the greedy algorithm regarding the fitness values on all artificial datasets except for two datasets $2b$, which proves that the greedy algorithm is able to find better solutions than two variant in most cases. The reason why RandEA gets better fitness values than SplitEA on two datasets $2b$ might be that the diversity introduction through splitting a random cluster into two clusters is not enough for those two datasets to find more clustering structure that gets better $U$, which can be reflected by the results of $U$. Following this line of diversity introduction, the reason why RandEA and SplitEA perform equally on Dataset 2a is that the diversity introduction in two EAs is somehow equal while the emphasis is different. In addition, SplitEA performs best regarding all metrics excepts for $Udelay$ on two Datasets $1c$. Similarly, the reason is that when SplitEA tries to cluster more points together, it might result in much delay. Besides, SplitEA gets best values regarding all metrics excepts for $K$ on two Datasets $3a$. The reason might be that the spitting of cluster in SplitEA causes solutions with better fitness values while inevitably increases the number of cluster.

\begin{table}[ht!]
\scriptsize
\begin{center}
\caption{The effect of problem-related parameters ($w$ and $\tau$) and EA-related parameters ($prob$, G and popsize) on the performance of SplitEA and two variants.}
\label{tab:ParaEAs}
\begin{tabular}{|c|ccc|ccc|ccc|}
\hline
w & \multicolumn{3}{|c|}{0.001} &\multicolumn{3}{|c|}{0.01} & \multicolumn{3}{c|}{1} \\
\hline
Algorithms & RandEA & CopyEA &	SplitEA &	RandEA & CopyEA &	SplitEA & RandEA & CopyEA &	SplitEA\\
\hline
K	&56.4863 	&\textbf{56.0373} 	&\textcolor{red}{53.2882} &	55.5353 	&\textbf{53.2020} 	&\textcolor{red}{52.0980} 	&55.3647 	&\textbf{52.3333} &	\textcolor{red}{51.7137} \\
U	&0.7758 &	\textbf{0.7712} &	\textcolor{red}{0.7623} 	&0.7788 	&\textbf{0.7687} &	\textcolor{red}{0.7659}	&0.7802 &	\textbf{0.7706} 	&\textcolor{red}{0.7684} \\
Udelay	&\textbf{0.0027}& 	\textcolor{red}{0.0020} &	\textbf{0.0026} 	&\textcolor{red}{0.0067} 	&\textbf{0.0069} 	&0.0077 	&\textcolor{red}{0.0079} 	&\textbf{0.0099} 	&\textbf{0.0100} \\
Uunder	&0.7730 	&\textbf{0.7691} 	&\textcolor{red}{0.7596} 	&0.7721 	&\textbf{0.7618} 	&\textcolor{red}{0.7582} 	&0.7723 &	\textbf{0.7607} &	\textcolor{red}{0.7583} \\
f&	0.8323 &	\textbf{0.8272} 	&\textcolor{red}{0.8156} 	&1.3342 	&\textbf{1.3007} 	&\textcolor{red}{1.2869} 	&56.1449 &	\textbf{53.1040} 	&\textcolor{red}{52.4821}\\
\hline
\hline
$\tau$ & \multicolumn{3}{|c|}{800} &\multicolumn{3}{|c|}{1000} & \multicolumn{3}{c|}{1500} \\
\hline
Algorithms & RandEA & CopyEA &	SplitEA &	RandEA & CopyEA &	SplitEA & RandEA & CopyEA &	SplitEA\\
\hline
K	&65.4824 	&\textbf{62.9980} &	\textcolor{red}{62.6255} 	&55.3647 	&\textbf{52.3333} 	&\textcolor{red}{51.7137} 	&40.5020 	&\textbf{36.9176} 	&\textcolor{red}{35.0765} \\
U	&0.8124 &	\textbf{0.8056} 	&\textcolor{red}{0.8048} &	0.7802 &	\textbf{0.7706} 	&\textcolor{red}{0.7684} 	&0.7061 	&\textbf{0.6806} 	&\textcolor{red}{0.6681} \\
Udelay	&\textcolor{red}{0.0058} &	\textbf{0.0063} &	0.0064 	&\textcolor{red}{0.0079} &	\textbf{0.0099} 	&0.0100 	&\textcolor{red}{0.0137} 	&\textbf{0.0169} 	&0.0185 \\
Uunder	&0.8066& 	\textbf{0.7993} &	\textcolor{red}{0.7983} &	0.7723 	&\textbf{0.7607} 	&\textcolor{red}{0.7583} 	&0.6924 	&\textbf{0.6637} 	&\textcolor{red}{0.6496} \\
f	&66.2947 &	\textbf{63.8037} &	\textcolor{red}{63.4303} 	&56.1449 &	\textbf{53.1040} &	\textcolor{red}{52.4821} 	&41.2081 	&\textbf{37.5982} &	\textcolor{red}{35.7446}\\
\hline
\hline
prob & \multicolumn{3}{|c|}{0.2} &\multicolumn{3}{|c|}{0.5} & \multicolumn{3}{c|}{0.8} \\
\hline
Algorithms & RandEA & CopyEA &	SplitEA &	RandEA & CopyEA &	SplitEA & RandEA & CopyEA &	SplitEA\\
\hline
K	&57.6667 	&\textbf{53.0333} &	\textcolor{red}{52.4118} &	55.3647 	&\textbf{52.3333} 	&\textcolor{red}{51.7137} &	57.6667 	&\textbf{53.0333} 	&\textcolor{red}{51.4843} \\
U	&0.7865 	&\textbf{0.7711} 	&\textcolor{red}{0.7684} 	&0.7802 	&\textbf{0.7706} 	&\textcolor{red}{0.7684} 	&0.7865 	&\textbf{0.7711} 	&\textcolor{red}{0.7673} \\
Udelay	&\textcolor{red}{0.0062} 	&0.0085 	&\textbf{0.0083} 	&\textcolor{red}{0.0079} 	&\textbf{0.0099} 	&\textbf{0.0100} 	&\textcolor{red}{0.0062} 	&\textbf{0.0085} 	&0.0101 \\
Uunder	&\textbf{0.7802} 	&\textcolor{red}{0.7626} 	&\textcolor{red}{0.7601} 	&0.7723 	&\textbf{0.7607} 	&\textcolor{red}{0.7583} 	&0.7802 	&\textbf{0.7626} 	&\textcolor{red}{0.7572} \\
f	&58.4531 	&\textbf{53.8045} &	\textcolor{red}{53.1801} 	&56.1449 	&\textbf{53.1040} 	&\textcolor{red}{52.4821} 	&58.4531 	&\textbf{53.8045} &	\textcolor{red}{52.2516}\\
\hline
\hline
G/popsize & \multicolumn{3}{|c|}{75/20} &\multicolumn{3}{|c|}{150/10} & \multicolumn{3}{c|}{300/5} \\
\hline
Algorithms & RandEA & CopyEA &	SplitEA &	RandEA & CopyEA &	SplitEA & RandEA & CopyEA &	SplitEA\\
\hline
K	&57.9608 	&\textbf{52.1157} 	&\textcolor{red}{51.9588} 	&55.3647 &	\textbf{52.3333} 	&\textcolor{red}{51.7137} 	&55.5961 	&\textbf{52.4314} &	\textcolor{red}{51.9451} \\
U	&0.7927 	&\textbf{0.7718} 	&\textcolor{red}{0.7697} 	&0.7802 	&\textbf{0.7706} 	&\textcolor{red}{0.7684} 	&0.7788 	&\textbf{0.7708} 	&\textcolor{red}{0.7680} \\
Udelay	&\textcolor{red}{0.0089} 	&0.0110 	&\textbf{0.0101} 	&\textcolor{red}{0.0079} 	&\textbf{0.0099} 	&\textbf{0.0100} 	&\textcolor{red}{0.0071} 	&0.0098 	&\textbf{0.0093} \\
Uunder	&0.7838 	&\textbf{0.7608} 	&\textcolor{red}{0.7596} 	&0.7723 	&\textbf{0.7607} 	&\textcolor{red}{0.7583} 	&0.7717 	&\textbf{0.7610} 	&\textcolor{red}{0.7587} \\
f	&58.7535 	&\textbf{52.8875} 	&\textcolor{red}{52.7285} &	56.1449 	&\textbf{53.1040} 	&\textcolor{red}{52.4821} 	&56.3749 	&\textbf{53.2021} 	&\textcolor{red}{52.7131}\\
\hline
\end{tabular}
\vspace{-0.48cm}
\end{center}
\begin{tablenotes}
      \footnotesize
      \item There are 30 independent runs. The values in this table are the mean value of the metrics under 30 runs. The best and the second best values obtained by the algorithm are highlighted in red and bold face, respectively. Friedman and Nemenyi statistical tests \cite{demvsar2006statistical} with the significance level 0.05 are used to indicate the statistical significance between compared algorithms. The metric value obtained by a given algorithm on one dataset is regarded as an observation to compose that algorithm’s group for the test, following Demsar’s guidelines \cite{demvsar2006statistical}. Therefore, there are 30 observations in each group for each metric on each parameter setting. If results obtained by two or three algorithms are highlighted with the same mark, it meas there is no significant difference between them.
    \end{tablenotes}
\end{table}

\paragraph{Fitness value curve of SplitEA and two variants on artificial datasets}

In order to have a better understanding of the proposed SplitEA and its two variants in the evolution process, the fitness value curve of three EAs on all artificial datasets in a randomly selected run is drawn and presented in figures \ref{fig.EAsCurve1} and \ref{fig.EAsCurve2}. As three EAs all initialize the population randomly at the first day, the curve of all days except for the first day is presented. It clear from those two figures that during the whole optimization process, the fitness vale of SplitEA is always better than that of RandEA and CopyEA on four datasets 3as and 1cs. As for the comparison results on datasets 1a and 2a, SplitEA is only worse than CopyEA on first day of all days. The reason might be that the spitting of clusters at the day cannot provide enough diversity as the RandEA gets the best on the first day. This case of insufficient diversity introduced by SplitEA becomes worse on two datasets 3as on which SplitEA performs best at initial generations while RandEA performs best at later generation of most days among all days.

\paragraph{The effect of different parameter settings on the performance of SplitEA and two variants}

In order to check the influence of different parameter settings on the performance of SplitEA and two variants, three algorithms are tested on the Milan Dataset which sets different values for two problem-related parameters $w$ and $\tau$ and three EA-related parameters $prob$, $G$ and $popsize$. The comparison results of SplitEA and two variants on dataset Milan under different parameter settings are presented in Table \ref{tab:ParaEAs}. It is clear from this table that under different settings of $w$, SplitEA significantly performs better than the greedy algorithm on all metrics and the fitness value. It is clear from this table that under different settings of all parameters, SplitEA gets best results, which shows that the proposed SplitEA is not sensitive with parameter setting.

\section{Conclusion and Future Work}
In this paper, we have presented a novel formulation for the computing resource allocation problem in O-RAN, so as to have a better representation of the problem, compared with the existing problem formulation. To this end, we manage to answer the first research question in Section \ref{sec:newProb}. Specifically, the proposed problem formulation equally considers the utilization rate of the computing resource and network delay in the fitness function. In addition, it removes the peak distribution, which is a needless criterion in the existing formulation. Afterwards, an evolutionary approach is designed tailored for solving the new problem formulation, which includes population initialization, mutation operator and random cluster splitting. Experimental studies have demonstrated that the proposed evolutionary algorithm significantly outperforms the greedy algorithm on all real-world datasets. It has also been proved that the proposed evolutionary algorithm is better than the greedy algorithm on most scenarios of different points distribution and different traffic pattern through testing them on several artificial dataset with different scenarios. In addition, computational studies have stated that the proposed approach's superiority remains even when adopting other parameter values. Similar experimental comparisons between the proposed evolutionary algorithm and two variants have also shown that random cluster splitting is important to achieve good fitness over time.

In the future, this work can be improved in several directions. Firstly, different size of computing resource can be considered in different clusters. Secondly, the communication delay between the BBU pool and base stations could be also taken into consideration. Lastly, more artificial intelligence techniques like prediction methods or optimization approaches can be leveraged to help the automation of O-RAN.

\section*{Acknowledgment}
This work has received funding from the European Union’s Horizon 2020 research and innovation programme under grant agreement number 766186. The work was also supported by the Program for Guangdong Introducing Innovative and Enterpreneurial Teams (Grant No. 2017ZT07X386), Shenzhen Peacock Plan (Grant No. KQTD2016112514355531) and the Program for University Key Laboratory of Guangdong Province (Grant No. 2017KSYS008).

\section*{References}

\bibliography{references}

\end{document}